\documentclass[a4paper,12pt]{article}
\usepackage[utf8]{inputenc}

\usepackage{
  amsmath,
  amsthm,
  bm,
  amssymb,
  hyperref,
  xcolor,
  caption,
  subcaption,
  graphicx,
  dsfont,
  multirow,
  booktabs,
  array,
  rotating,
  enumitem,
  mathtools,
  placeins
}

\hypersetup{
	hidelinks,
  colorlinks = true,
  linkbordercolor = {white},
  citecolor = blue,
  urlcolor = blue,
  linkcolor = blue,
  linktoc = all               
}

\usepackage[
  top=2cm,
  bottom=2cm,
  left=1in,
  right=1in,
  headheight=17pt,  
  includehead,
  includefoot,
  heightrounded,    
]{geometry}

\usepackage{natbib}
\setcitestyle{authoryear}
\theoremstyle{plain}

\theoremstyle{definition}

\usepackage{amsfonts}
\usepackage{comment}
\usepackage{textcomp}             
\usepackage{algorithm}            
\usepackage[noend]{algpseudocode} 

\newcommand{\boldtitle}[1]{\title{\textbf{\boldmath{#1}}}}
\newcommand{\keywords}[1]{\medskip \par \noindent \textbf{Keywords}\quad #1}
\newcommand{\mat}[1]{\boldsymbol{#1}}   
\newcommand{\vect}[1]{\boldsymbol{#1}}  

\newcommand{\pkg}[1]{\texttt{#1}}
\newcommand{\proglang}[1]{\textsf{#1}}

\DeclareMathOperator*{\argmin}{arg\,min}
\DeclareMathOperator{\median}{median}

\boldtitle{Towards Reliable Recommender Systems for Rating Data}

\author{
  Aurore Archimbaud\footnote{Parts of this research were done while Aurore Archimbaud was at Erasmus University Rotterdam.} \\
  TBS Business School \\
  \href{mailto:a.archimbaud@tbs-education.fr}{\texttt{a.archimbaud@tbs-education.fr}}
  \and
  Andreas Alfons \\
  Erasmus University Rotterdam \\
  \href{mailto:alfons@ese.eur.nl}{\texttt{alfons@ese.eur.nl}}
  \and
  Ines Wilms \\
  Maastricht University \\
  \href{mailto:i.wilms@maastrichtuniversity.nl}{\texttt{i.wilms@maastrichtuniversity.nl}}
}

\date{}

\begin{document}

\maketitle

\begin{abstract}
\noindent
Recommender systems are widely used in the digital landscape to match users with content fitting their preferences. However, growing concerns about fake accounts, strategic manipulation, and other deceptive online behavior place increasing pressure on the reliability of these systems.
A common statistical approach behind recommender systems is so-called matrix completion, which predicts how users would rate items they have not yet consumed based on patterns in observed ratings.
Realistically applying matrix completion methods requires jointly addressing several overlooked challenges:
(i)~ratings on discrete scales (such as 1--5 stars);
(ii)~the presence of malicious users who deliberately manipulate the system to their advantage through fake profiles;
(iii)~ratings missing not at random since users are more likely to consume items they expect to like;
and (iv)~fostering transparency, reproducibility, and stability.
We jointly address these challenges by proposing a novel method, \emph{Robust Discrete Matrix Completion} (RDMC), designed to capture the key characteristics of sparse rating data while remaining reliable in the presence of manipulation.
We evaluate RDMC through two case studies and carefully designed simulation experiments.
Our work thereby offers a statistically-sound blueprint for future studies on how to evaluate recommender systems under realistic scenarios.

\end{abstract}

\vfill

\paragraph{Funding information}
This work is supported by grants of the Dutch Research Council (NWO), research program Vidi, grant numbers VI.Vidi.195.141 (Andreas Alfons) and VI.Vidi.211.032 (Ines Wilms).

\newpage


\section{Introduction}
\label{sec:intro}

Recommender systems play a key role in digital platforms, particularly  online retail and advertising, helping users discover products efficiently while driving substantial commercial value \citep{leblanc2023}.
Their reliability is a societal and economic concern, as these algorithms increasingly shape consumer decisions, content exposure, and market dynamics.

Recommender systems are, however, highly susceptible to adversarial manipulation through attacks with fake user profiles that deliberately manipulate recommendations to their advantage \citep[e.g.,][]{van2010manipulation, nguyen_manipulating_2024}, creating a market for fake and strategic reviews and accounts (e.g., \citealp{he2022market, adamopoulos2024spillover, huang2025economics}).
These manipulations undermine the integrity of recommender systems and erode user trust.
In response, regulatory authorities worldwide such as  the \citeauthor{FTC2019} (FTC; \citeyear{FTC2019}), the \citeauthor{CMA2021} (CMA; \citeyear{CMA2021}), and the \citet{DSA2022} have imposed consent orders and million-dollar penalties targeting deceptive practices, illustrating the issue's real-world scale and urgency (e.g., FTC \citeyear{FTC_FashionNova_2022, ftc2024rytr}, \citealp{icpen2025news1440}). They have also introduced measures to safeguard algorithmic integrity and protect consumers urging digital platforms to proactively detect and address fake profiles and suspicious behavioral patterns, including those generated by artificial intelligence. A recent example is Amazon's agreement to CMA undertakings to curb fake reviews on its platform (CMA \citeyear{CMA_Amazon_2025}).

While academic research on recommender systems has flourished, the crucial challenge of ensuring \emph{reliable} recommender systems remains comparatively underexplored.
We define reliability as encompassing both robustness to adversarial manipulations and stability of predictions to algorithmic choices.
This creates a  disconnect and growing tension between methodological advances in academia on the one hand  and the much needed reliability in practice on the other hand.
We address that tension by designing, applying, and empirically evaluating a new, reliable matrix completion method called \emph{Robust Discrete Matrix Completion (RDMC)} that captures the key characteristics of online ratings in a statistically-sound manner.

From a statistical perspective, recommender systems are often framed as matrix completion problems for a sparse user-item rating matrix (e.g., ranging from one star to five stars), with predictions being used to recommend new items to users. Early work on matrix completion dates back to, e.g., \citet{Achlioptas2001} and \citet{srebro2004maximum}.
It has been actively studied since \citet{candes2009} and recent overviews are provided by, e.g., \citet{leblanc2023}, \citet{gheewala2025depth} and \citet{ raza2026comprehensive}.

We contribute to the literature by addressing four persistent gaps that create a tension between  academic research and practical relevance in recommender systems.

\textbf{1.~Discreteness of Rating Data.}
The majority of existing matrix completion approaches are formulated over the real number domain and produce continuous predictions, despite recommender systems
applications commonly featuring discrete and bounded rating data.
Exceptions that include discreteness or box constraints include \citet{huang2013robust}, \citet{tatsukawa2018box}, and more recently \citet{bertsimas2023interpretable}.
While observed ratings may be interpreted as discrete measurements of a latent continuous sentiment, the underlying continuous distributions are not identified without additional assumptions \citep[cf.][]{bond2019identification}, rendering comparisons of predictions across items invalid.
We guarantee the statistical soundness of RDMC for discrete ratings by restricting the predictions to the given rating scale via a discreteness constraint (Section~\ref{sec:method}).

\textbf{2.~Robustness in Presence of Attacks.}
Increasingly many sophisticated rating prediction algorithms exist \citep[e.g.,][]{gheewala2025depth, raza2026comprehensive}. However, less attention has been paid to their robustness in presence of corrupted/manipulated observations (\emph{outliers}) such as fake profiles, despite their practical prevalence.
Work on robust matrix completion includes robust nuclear norm minimization \citep[e.g.,][]{huang2013robust, elsener2018robust} or robust matrix factorization \citep[e.g.,][]{tang2020robust, wang2024robust}.
We further narrow this gap by responding to the research call in \citet{gheewala2025depth} to develop methods that remain robust under adversarial manipulation.
RDMC fulfills this via a built-in robust loss function (Section~\ref{sec:method}).

\textbf{3.~Impact of Missing Data Mechanisms.}
The effect of missing data mechanisms other than missing completely at random (MCAR) on matrix completion is not well understood; notable exceptions being \citet{mao2019matrix} for missing at random (MAR) as well as \citet{choi2024matrix} and \citet{xu2025learning} for missing not at random (MNAR).
Yet, in practice, missingness in recommender systems is inherently preference-dependent \citep{leblanc2023}:
Users predominantly consume and then rate items they expected to like, intrinsically linking the probability of missingness to their preferences.
We study the performance of RDMC under realistic MNAR settings to ensure its validity under practical conditions.

\textbf{4.~Transparency, Reproducibility, and Stability in Recommender Systems.}
Transparency and reproducibility are key concerns in the current scientific landscape on recommender systems \citep{leblanc2023}. \cite{ferrari_dacrema_troubling_2021} highlight a frequent lack of missing methodological details (especially on hyperparameter optimization), insufficient baseline method tuning and the absence of standardized benchmarks.
To enhance reproducibility and generalizability, we introduce a transparent pipeline for designing, applying, and comparing matrix completion methods under adversarial manipulation and alternative algorithmic choices, focusing respectively on robustness and stability of rating predictions.
We consider two empirical case studies, with motivating datasets and associated scientific questions presented in Section~\ref{sec:data-and-RQ} and data analysis in Section~\ref{sec:applications}, as well as simulations in Section~\ref{sec:simulations}.
The latter are carefully designed to be more realistic than previous studies, thus providing a  conceptual contribution: Our experiments provide statistically-sound, holistic blueprints for future studies on how to design simulation experiments for recommender systems, and how to realistically apply and compare methods.
We provide implementations of the methodology in package \pkg{RMCLab} \citep{RMCLab} for the statistical computing environment \proglang{R} \citep{Rcoreteam}, available from \url{https://CRAN.R-project.org/package=RMCLab}.
\emph{Replication files of all analyses will be made publicly available upon acceptance of this manuscript.}

RDMC provides the first joint treatment of the four above described tensions.
We assess RDMC  against widely-used procedures including Soft-Impute \citep{mazumder2010spectral, hastie2015matrix} and neural matrix factorization \citep{he2017neural}.
Our study  hereby contributes to current empirical practice in recommender systems in the following ways:
we (i) adapt existing methods to better account for the discrete nature of rating data, (ii)  extend neural matrix factorization to explicit-feedback settings by making full use of the information contained in discrete ratings, and (iii) carefully tune all baseline methods to ensure fair and meaningful comparisons.
We find that RDMC is \emph{reliable}: It captures key characteristics of rating data with built-in \emph{robustness} to adversarial
manipulation and offers \emph{stable} rating predictions with respect to algorithmic choices.


\section{Motivating datasets and scientific questions}
\label{sec:data-and-RQ}
We analyze rating data from two popular data sources in the context of recommender systems: MovieLens and Yahoo!~Music.

\begin{figure}[t]
    \centering
    \includegraphics[width=\textwidth]{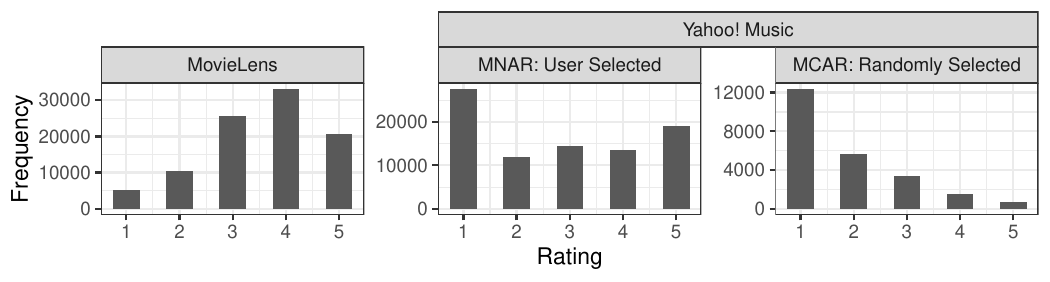}
    \caption{Overall distribution of observed ratings in the MovieLens 100K data (left) and the Yahoo! Music datasets (middle and right).}
    \label{fig:movielens_yahoo_ratings}
\end{figure}

{\bf MovieLens.}
The famous MovieLens  dataset \citep{harper2015movielens} is extensively analyzed in previous studies on recommender systems as documented by recent review articles \citep[e.g.,][]{gheewala2025depth, raza2026comprehensive}.
We use the MovieLens 100K dataset consisting of 100,000 movie ratings on a scale from 1 to 5, of $n = 943$ users on 1,682 movies.
We restrict our analysis to the $p = 939$ movies that are rated at least 20 times.
The observed ratings are left-skewed with a majority of medium to high ratings, see  Figure~\ref{fig:movielens_yahoo_ratings} (left).

Since the MovieLens 100K dataset is widely used, it forms a natural baseline for evaluating matrix completion methods under controlled adversarial conditions.
The goal of this analysis is thus to study robustness to adversarial manipulation via injected fake user profiles.
We inject nuke attacks, a well-known attack strategy aimed at demoting certain items \citep{mobasher2007toward} (see Section \ref{sec:attacks} for further details).
The choice of the 100K dataset ensures computational feasibility of our extensive empirical study, including multiple attack scenarios, repeated analyses across different data splits for training and evaluation, and a range of method configurations.

{\bf Yahoo!~Music.}
The Yahoo!~Music data contain user–item interactions in the form of ratings that users assign to songs.
Our goal is to investigate how different missing data mechanisms affect the performance of matrix completion methods.
To this end, we use two related Yahoo!\ Music datasets that provide an experimental framework for this comparison.
In the first dataset (\emph{User Selected}), the users choose the songs they want to rate.
In the second dataset (\emph{Randomly Selected}), 10 songs from a pool of $p_{2} = 1,000$ are randomly assigned to them. Hence, the former constitutes an MNAR mechanism, the latter an MCAR mechanism.
To ensure a fair comparison across both, we restrict our analysis to the $n = 2,361$ users who are present in both datasets and rated at least 20 songs in the first dataset.
We further limit the first dataset to the $p_{1} = 921$ songs with at least 20 ratings.  In both datasets, we have more than 95\% of missing values (96\% for MNAR, 99\% for MCAR).

The missing data mechanism leads to different rating distributions, see Figure~\ref{fig:movielens_yahoo_ratings} (middle and right). When the users can choose the songs to evaluate (MNAR, middle), the distribution is almost bimodal with 1-star ratings being the most frequent, followed by 5-star ratings. When the songs to be rated are randomly assigned (MCAR, right), the distribution is highly right-skewed with decreasing frequency for increasing ratings, and very few high ratings.

The MovieLens and Yahoo!~Music datasets share two key features that are central to our analysis. First, ratings are inherently discrete (on a 1--5 scale) and highly skewed.
These fundamental data features should be explicitly accounted for in a statistically sound manner.
While a prevalent business practice is to identify and remove fraudulent reviews or malicious user profiles before applying recommendation algorithms,   \cite{adamopoulos2024spillover} cautions against this, showing that it can result in persistent, long-term negative effects. Rating values cannot be easily interpreted as corrupted or manipulated observations and standard outlier detection methods, e.g., based on continuous location–scale assumptions are not directly applicable to discrete data. These may inadvertently discard an entire `extreme' rating category (like 1- or 5-star ratings) to be outlying, thereby distorting the underlying distribution.
This  instead calls for a robust method \citep[e.g.,][]{avella2015} that offers built-in protection against corrupted or manipulated profiles during estimation to accommodate the presence of outliers without compromising performance.

Second, the considered datasets represent explicit feedback settings in recommender systems, where users directly express preferences (e.g., \citealp{leblanc2023}). This contrasts with implicit feedback, where preferences are inferred from user behavior such as clicks or views. While explicit feedback is typically more costly to collect, it provides richer and more informative signals about user preferences.
Despite this, much of the recent matrix completion and deep learning literature focuses on methods designed for implicit feedback, while still relying on rating datasets like MovieLens for benchmarking. Explicit ratings are then often converted into implicit signals, resulting in information loss and underutilization of the observed data structure.
We instead focus on discrete rating prediction  under explicit feedback, directly modeling ratings to fully exploit the information they contain.

We use these datasets as complementary settings for evaluating how challenging features of rating data in recommender systems influence the performance of matrix completion methods. Furthermore, we assess the stability of predictions via sensitivity analyses on algorithmic choices.
We specifically address the following scientific questions:
\begin{description}[labelwidth = 0.75cm, leftmargin = !]
\item[Q1] How can matrix completion methods
be made statistically sound for explicit-feedback data with discrete ratings?
\item[Q2] How robust are matrix completion methods to adversarial manipulation of rating data?
\item[Q3] How does the missingness mechanism in rating data affect the performance of matrix completion methods?
\item[Q4] How can we design realistic simulation blueprints for evaluating matrix completion methods that capture key characteristics of rating data?
\end{description}
We tackle these questions by introducing a novel methodological pipeline that designs, applies, and evaluates a new matrix completion method, RDMC, which we introduce next.


\section{A robust procedure for discrete matrix completion}
\label{sec:method}

To address scientific question Q1, we start by formulating our optimization problem in Section~\ref{sec:problem} and presenting the corresponding algorithm in Section~\ref{sec:algorithm}. Section~\ref{sec:centering} provides further algorithmic details, Section~\ref{sec:loss} introduces different choices of robust loss functions for RDMC, and Section~\ref{sec:validation} discusses regularization parameter selection.
Sections~\ref{sec:benchmarks} and \ref{sec:evaluation} detail the benchmark methods and evaluation metrics we use throughout the paper.

\subsection{Problem formulation}
\label{sec:problem}

Suppose that we observe an incomplete rating matrix $\mat{R}$ of $n$ rows (the individuals providing the ratings) and $p$ columns (the rated items) with elements $ R_{ij} \in \{1, 2, \dots, K\}$. For practical reasons (see Section~\ref{sec:centering} for further discussion), we consider a column-centered matrix $\mat{X}$ with elements $X_{ij} \in \mathcal{C}_{j} = \{c_{1}^{j}, \dots, c_{K}^{j}\} \subset \mathbb{R}$, where $c_{1}^{j} < \dots < c_{K}^{j}$ for $j = 1, \dots, p$. Note that the values of the rating categories may vary between columns of~$\mat{X}$. Hence, the assumption that the original rating matrix $\mat{R}$ takes values in $\{1, 2, \dots, K\}$ is purely for simplicity and without loss of generality, and the proposed procedure extends in a straightforward manner to settings where even the number of rating categories may differ per column.

Let  $\Omega$ denote the index set of observed entries in $\mat{X}$ and define the projection $P_{\Omega}$ to be the $(n \times p)$-dimensional matrix whose elements are given by
\begin{equation*}
\left( P_\Omega(\mat{X}) \right)_{ij} =
\begin{cases}
  X_{ij} & \text{ if } (i,j) \in \Omega, \\
  0 & \text{ otherwise.}
\end{cases}
\end{equation*}
Matrix completion is often based on minimizing the mean squared error for the observed elements subject to a nuclear norm constraint as a relaxation of a low-rank constraint. Specifically, the widely-used procedure Soft-Impute \citep{mazumder2010spectral, hastie2015matrix} solves the optimization problem (in Lagrangian form)
\begin{equation}
\label{eq:SI}
\min_{\mat{L}}
\frac{1}{2} \| P_{\Omega}(\mat{X}) - P_{\Omega}(\mat{L}) \|_{F}^{2} +
\lambda \| \mat{L} \|_{*},
\end{equation}
where $\| \cdot \|_{F}$ denotes the Frobenius norm and $\| \cdot \|_{*}$ the nuclear norm, while $\lambda \geq 0$ is a regularization parameter.

In order to preserve the discrete nature of the ratings, one could add a discreteness constraint to~\eqref{eq:SI} \citep[cf.][]{huang2013robust}. However, requiring the elements of $\mat{L}$ to be discrete may not yield a solution, since such a discreteness constraint and a rank constraint are unlikely to be fulfilled simultaneously. To resolve this issue, an ancillary continuous matrix~$\mat{Z}$ can be introduced into problem \eqref{eq:SI} such that the discreteness constraint operates on $\mat{L}$ and the nuclear norm constraint on $\mat{Z}$, while ensuring that $\mat{L}$ and $\mat{Z}$ remain as close as possible.
In addition, the squared Frobenius norm in~\eqref{eq:SI} does not protect against corrupted observations (such as fake profiles in recommender systems).
Indeed, this norm penalizes large errors quadratically, implying that corrupted observations with large errors may dominate the loss function and significantly influence the recommender system.
For increased protection, we replace it with a robust loss function that puts less emphasis on large errors,
thereby preventing them from disproportionately affecting the recommender system. More specifically, we replace it with a (pseudo-)norm $\| \mat{Y} \|_{\rho} = \sum_{i,j} \rho(Y_{ij})$ based on a robust loss function $\rho$ \citep[cf.][]{tang2020robust}.
Putting all of this together, we obtain the following optimization problem in augmented Lagrangian form:
\begin{equation}
\label{eq:RDMC}
\begin{array}{r l}
{\displaystyle \min_{\mat{L}, \mat{Z}}} &
{\displaystyle \| P_{\Omega}(\mat{X}) - P_{\Omega}(\mat{L}) \|_{\rho} +
\lambda || \mat{Z}||_{*} +
\langle \mat{\Theta}, \mat{L} - \mat{Z} \rangle_{F} +
\frac{\mu}{2} || \mat{L} - \mat{Z}||_{F}^{2}} \\
\text{subject to} & L_{ij} \in \mathcal{C}_{j}, \quad i = 1, \dots, n, j = 1, \dots, p,
\end{array}
\end{equation}
where $\langle \cdot,\cdot \rangle_{F}$ denotes the Frobenius inner product, $\mat{\Theta}$ is a multiplier adjusting for the discrepancy between $\mat{L}$ and $\mat{Z}$, and $\mu$ is an additional regularization parameter. It should be noted that~\eqref{eq:RDMC} generalizes the optimization problem formulated in \citet{huang2013robust} to a wider class of robust loss functions and by allowing different values of the rating categories between columns of the (column-centered) rating matrix.

\subsection{Algorithm}
\label{sec:algorithm}

The formulation of the optimization problem~\eqref{eq:RDMC} lends itself to an alternating direction method of multipliers (ADMM) algorithm \citep{boyd2011distributed}, which follows along similar lines as that of \citet{huang2013robust}. For a given value of the regularization parameter~$\lambda$, the following steps are iterated until convergence (after initialization as described in Section~\ref{sec:centering}).

First, consider $\mat{L}$ fixed and solve~\eqref{eq:RDMC} for $\mat{Z}$. Combining the terms for the Frobenius inner product and Frobenius norm, dropping constant terms, and division by $\mu$ yields the equivalent minimization problem
\begin{equation*}
\min_{\mat{Z}}
{\textstyle
\frac{1}{2} \| (\mat{L} + \frac{1}{\mu} \mat{\Theta}) - \mat{Z} \|_{F}^{2} +
\frac{\lambda}{\mu} \| \mat{Z} \|_{*}
}.
\end{equation*}
The solution is given by the soft-thresholded singular value decomposition (SVD) of $\mat{L} + \frac{1}{\mu} \mat{\Theta}$ \citep{cai2010singular}. That is, with $\mat{L} + \frac{1}{\mu} \mat{\Theta} = \mat{U} \mat{D} \mat{V}^{\top}$, where $\mat{U}$ is $(n \times q)$-dimensional, $\mat{V}$ is $(p \times q)$-dimensional, $\mat{D}$ is a $(q \times q)$-dimensional diagonal matrix whose diagonal elements are denoted by $d_{1}, \dots, d_{q}$, and $q \leq \min(n, p)$ denoting the rank, we obtain
\begin{equation}
\mat{Z} = \mat{U} S(\mat{D}) \mat{V}^{\top}, \label{eq:Zupdate}
\end{equation}
where $S(\mat{D}) = \text{diag} \left( (d_{1} - \frac{\lambda}{\mu})_{+}, \dots, (d_{q} - \frac{\lambda}{\mu})_{+} \right)$ with $(y)_{+} = \max(y, 0)$.

Second, consider $\mat{Z}$ fixed and solve~\eqref{eq:RDMC} for $\mat{L}$. Using similar operations as described above but without rescaling, we arrive at the equivalent minimization problem
\begin{equation*}
\begin{array}{r l}
{\displaystyle \min_{\mat{L}}} &
\| P_{\Omega}(\mat{X}) - P_{\Omega}(\mat{L}) \|_{\rho} +
\frac{\mu}{2} \| \mat{L} - \mat{Z} + \frac{1}{\mu} \mat{\Theta} \|_{F}^{2} \\
\text{subject to} & L_{ij} \in \mathcal{C}_{j}, \quad i = 1, \dots, n, j = 1, \dots, p.
\end{array}
\end{equation*}
This can be solved element-wise, with the solution being given by
\begin{equation}
L_{ij} =
\begin{cases}
{\displaystyle \argmin_{c_{k} \in \mathcal{C}_{j}}} \
\rho(c_{k}- X_{ij}) + \frac{\mu}{2} (c_{k} - Z_{ij} + \frac{1}{\mu} \Theta_{ij})^2 &
\text{ for } (i,j) \in \Omega, \\
{\displaystyle \argmin_{c_{k} \in \mathcal{C}_{j}}} \
(c_{k} - Z_{ij} + \frac{1}{\mu} \Theta_{ij})^2 &
\text{ for } (i,j) \notin \Omega.
\end{cases} \label{eq:updateL}
\end{equation}

Third, update the discrepancy parameter $\mat{\Theta} \leftarrow \mat{\Theta} + \mu (\mat{L} - \mat{Z})$ and the regularization parameter $\mu \leftarrow \delta \mu$ with $\delta > 1$, so that $\mu$ grows exponentially in the number of iterations to speed up convergence.

Pseudo-code is presented in Algorithm \ref{algorithm-rdmc} of Appendix~\ref{app:algorithm}.
Following \citet{huang2013robust} and \citet{tang2020robust}, we initialize $\mu = 0.1$ and set $\delta = 1.05$.  As convergence criterion, we take the relative change in the objective function from~\eqref{eq:RDMC} falling below a given threshold  $\varepsilon_\text{tol} = 10^{-4}$ and set the maximum number of iterations to $t_\text{max}=100$, as  this sufficed for convergence in all our numerical experiments.

\subsection{Column centering and initialization}
\label{sec:centering}

Since the algorithm contains a soft-thresholded SVD step, it is important that the observed (incomplete) rating matrix $\mat{R}$ is centered. One possibility is to center each column by the midpoint of the rating scale (i.e., the mean of the minimum and maximum rating category).
While this can be expected to work well if the rating distributions of the columns are symmetric around the midpoint, such a setting is unrealistic in practical applications. In recommender systems, the distributions of popular items, for instance, are typically skewed towards higher ratings, with the maximum rating often being the most frequent.
Hence, we center the $j$th column in the given data matrix $\mat{R}$ by the median $M_j$ of its observed cells, i.e., we obtain $\mat{X}$ by setting $X_{ij} = R_{ij} - M_{j}$, $i = 1, \dots, n$, $j = 1, \dots, p$. Accordingly, we transform the set of rating categories to $C_{j} = \{c_{1}^{j}, \dots, c_{K}^{j}\}$ with $c_{k}^{j} = k - M_{j}$ for $k = 1, \dots, K$. This highlights the need for a procedure that allows for different rating categories in different columns, a feature that our proposed procedure RDMC accommodates.

Using the median-centered matrix $\mat{X}$, we initialize the matrix $\mat{L} = P_{\Omega}(\mat{X})$ (corresponding to median imputation) and the discrepancy parameter $\mat{\Theta}$ as an $(n \times p)$-dimensional matrix of zeros. However, the algorithm is typically applied for a grid of values for~$\lambda$ (see Section~\ref{sec:validation}). Then the obtained solutions for $\mat{L}$ and $\mat{\Theta}$ for a given value of  $\lambda$ are used as starting values for the next value of $\lambda$. As $\mat{\Theta}$ adjusts for the discrepancy between $\mat{L}$ and $\mat{Z}$ (which should increase with increasing $\lambda$), the values of $\lambda$ should thereby be sorted in ascending order.

\subsection{Loss functions}
\label{sec:loss}

In order to reduce the influence of large errors due to corrupted observations such as fake profiles, we consider the following robust loss functions for the (pseudo-)norm~$\|\cdot\|_{\rho}$ in~\eqref{eq:RDMC}, which are common choices in the literature on robust methods:
\begin{itemize}
    \item \emph{pHuber}: The pseudo-Huber loss $\rho(y) = \tau^2 (\sqrt{1 + (y / \tau)^2} - 1)$ with parameter $\tau$. We include this loss due to its successful application to rating-scale data in a different context, namely autoencoder neural networks for detecting careless responding in surveys from the behavioral sciences \citep{alfons2024open, welz2024carelessonset}. We link the parameter $\tau$ to the step size between rating categories by setting $\tau = 1$.
    \item \emph{absolute}: The absolute loss $\rho(y) = |y|$.
    \item \emph{truncated}: A truncated variant of the absolute loss given by $\rho(y) = \min (|y|, \tau)$. We set $\tau = (K - 1) / 2$, i.e., the loss is truncated at half the range of the rating categories.
\end{itemize}
Since the update step for $\mat{L}$ yields separable problems for its elements (see Section~\ref{sec:algorithm}), the use of a nonconvex function such as the truncated absolute loss comes at no cost to computational complexity. A nonconvex loss may yield greater robustness in selecting the regularization parameter $\lambda$ when minimizing the loss on a test set, which is discussed next.

\subsection{Selection of the regularization parameter}
\label{sec:validation}

To select the regularization parameter $\lambda$ from a grid of candidate values based on out-of-sample prediction performance, some of the observed elements of $\mat{R}$ can be set to missing values for subsequent use as a validation set. For measuring the prediction error on the validation set, a natural choice is to apply the same loss function $\rho$ that is used for fitting the algorithm on the training set.
It is possible to apply a cross-validation scheme whereby the observed elements are randomly divided into blocks, with each block being used as validation set once. However, even if the rows are independent, elements within the same row are not. It is therefore unclear if cross-validation would reduce the correlations among prediction errors on the different validation sets, compared to repeated holdout validation whereby a proportion of the observed elements are randomly selected in each replication to form the validation set. Hence, we prefer repeated holdout validation, as the proportion of observations in the validation set and the number of replications can be chosen independently.

Throughout, we tune RDMC via repeated holdout validation with 5 replications and 10\% of the observed cells to be randomly selected as validation set. The candidate values for $\lambda$ are given by a logarithmic grid of ten values between 0.01 and 1, which are scaled by the largest singular value of $P_{\Omega}(\mat{X}_{\text{train}})$, with $\mat{X}_{\text{train}}$ denoting the median-centered training data.

\subsection{Benchmark methods}
\label{sec:benchmarks}

We compare \emph{RDMC} against several benchmark methods. Following recommendations of \cite{ferrari_dacrema_troubling_2021}---which are motivated by the fact that even simple methods have been shown to outperform complex neural networks in some settings---these benchmarks are selected from different methodological families ranging from basic and advanced statistical approaches to recent deep learning techniques.
To ensure a fair and meaningful comparison, we also adapt the benchmark methods to better account for the discrete nature of rating data and we carefully tune their hyperparameters.

We consider \emph{median} imputation with a discretized variant \emph{median-discretized} and \emph{mode} imputation as basic methods.
Among advanced statistical methods, we consider Soft-Impute (\emph{SI}) \citep{mazumder2010spectral, hastie2015matrix} with a discretized variant \emph{SI-discretized}. Finally, we consider neural matrix factorization (\emph{NeuMF}) \citep{he2017neural}, which we adapted for explicit feedback, with a discretized variant \emph{NeuMF-discretized}.
Appendix~\ref{app:benchmarks} provides detailed descriptions, including hyperparameter selection.

\subsection{Evaluation metrics}
\label{sec:evaluation}

In the absence of adversarial manupulation, the methods are evaluated using the mean absolute error (MAE) of predictions on an out-of-sample test set, i.e.,
\begin{equation}
\label{eq:MAE}
MAE = \frac{1}{|\Gamma_{\text{test}}|} \sum_{(i,j) \in \Gamma_{\text{test}}} |R_{ij} - \widehat{R}_{ij}|,
\end{equation}
where $\Gamma_{\text{test}}$ contains the indices of cells in $\mat{R}$ that form the test set, and $\widehat{R}_{ij}$ is the prediction of the cell~$R_{ij}$. Note that we favor the MAE over the more commonly used mean squared error (MSE) due to the discrete nature of the ratings.

Under adversarial manipulation, we follow \cite{mobasher2007toward} and compute the mean prediction shift (MPS) in the variable that is the target of a fake profile attack, i.e.,
\begin{equation*}
MPS =
\frac{1}{|\Gamma_{\text{target}}|} \sum_{i \in \Gamma_{\text{target}}}
\left (\widehat{R}_{ij_{\text{target}}}^{\text{after}} - \widehat{R}_{ij_{\text{target}}}^{\text{before}} \right),
\end{equation*}
where $\Gamma_{\text{target}}$ is the index set of missing cells in the target variable $j_{\text{target}}$ of $\mat{R}$, while $\widehat{R}_{ij}^{\text{before}}$ and  $\widehat{R}_{ij}^{\text{after}}$ denote the prediction of the cell~$R_{ij}$ before and after the attack, respectively; see Section~\ref{sec:attacks} for details on the attacks.


\section{Empirical case studies}
\label{sec:applications}

We now analyze in detail the datasets outlined in Section \ref{sec:data-and-RQ} in order to address scientific questions Q2 and Q3. Throughout this section, we further analyze the sensitivity of our findings with respect to the following two choices.

\textbf{Data splits.} We perform repeated analyses on 10 different data splits to assess the stability of the methods when trained and evaluated on different data. For NeuMF, the repeated analyses also imply that different hyperparameter combinations are sampled in the validation procedure. In addition to the average evaluation metrics across the data splits, we report the respective minimum and maximum values.

\textbf{Stopping criterion.}
Since computational burden is an important practical consideration in recommender systems, we assess the stability of the methods with respect to their stopping criteria. That is, we investigate whether reducing the computational burden comes at a cost of performance. The standard convergence threshold for RDMC and SI of $10^{-4}$ is denoted as the \emph{strict} criterion. For RDMC, the \emph{liberal} criterion corresponds to stopping after a maximum of 10 iterations. For SI, we follow the strategy of \citet{hastie2015matrix} by setting a larger convergence threshold of $10^{-3}$ for the \emph{liberal} criterion. Note that such a sensitivity analysis is not (easily) applicable for NeuMF, as deep learning frameworks commonly incorporate built-in early-stopping criteria during training.

\subsection{MovieLens data: The impact of attacks}
\label{sec:MovieLens}

We use the MovieLens 100K data to investigate the robustness of the matrix completion methods to adversarial manipulation, thereby addressing Q2.
In Section  \ref{sec:attacks}, we detail how we inject fake profile attacks.
We then establish baseline prediction results in the absence of attacks in Section~\ref{sec:movielens-baseline}, and present the main
results regarding Q2 in Section~\ref{sec:movielens-attacks}.

\subsubsection{Attacks}
\label{sec:attacks}

Concerning adversarial manipulation of recommender systems, \citet{mobasher2007toward} introduced a taxonomy and formal framework for profile injection attacks (also known as shilling attacks), which we adopt.
We thereby focus on so-called nuke attacks aimed at demoting a certain \emph{target} item, i.e., decreasing the probability of it being recommended.
We apply three efficient attack schemes, denoted by \emph{average}, \emph{reverse bandwagon}, and \emph{love/hate}.

As nuke attacks are intended for popular items, we select as the target item the one with the highest sample mean of the observed ratings among all items with at least 50 ratings. Fake profiles then assign the minimum rating 1 to the target item.
The number of fake profiles, which we also refer to as the attack size, is thereby determined as a proportion $\varepsilon \in \{0.10, 0.15, 0.20\}$ of the number of observed ratings in the target item.

Table~\ref{tab:attacks_desc} summarizes the strategy in the three attack schemes regarding so-called \emph{selected items} (which are chosen by the attacker based on specific characteristics) and \emph{filler items} (which are randomly chosen).
In the \emph{average} attack, filler items are assigned ratings based on the item mode. The idea is that the fake profiles have typical ratings in other items but dislike the target item. We thereby use the item mode instead of the item average, following the recommendation of \cite{turk_robustness_2019} for discrete ratings.
The \emph{reverse bandwagon} attack is intended to associate the target item with disliked items. Hence, it uses unpopular items as selected items and assigns the minimum rating to both the selected and filler items.
As unpopular items, we take the items with the lowest average observed ratings, provided that at least 20 ratings are observed.
In the \emph{love/hate} attack, the filler items receive the maximum rating so that the fake profiles love any other item while hating the target item.

\begin{table}[b]
    \caption{Overview of the selected nuke attack schemes.}
    \centering
    \setlength{\tabcolsep}{4pt}
    \begin{tabular}{lclccclcc} \hline
         & & \multicolumn{3}{c}{Selected items} & & \multicolumn{3}{c}{Filler items}  \\
         \cline{3-5} \cline{7-9}
         Attack type & & Selection & Fraction & Rating & & Selection & Fraction & Rating  \\
         \hline
         Average & & Not used & & & & Random & 0.1 & Item mode \\
         Reverse bandwagon & & Unpopular & 0.1 & $1$ & & Random  & 0.1 & $1$ \\
         Love/hate & & Not used & & & & Random & 0.1 & $K$  \\ \hline
    \end{tabular}
    \label{tab:attacks_desc}
\end{table}

\subsubsection{Results on baseline prediction performance}
\label{sec:movielens-baseline}

We assess the baseline prediction performance of RDMC and the benchmark methods in the absence of attacks. To this end, we set aside 20\% of the observed ratings as a test set, and use the remaining ratings for training and hyperparameter validation.

\begin{figure}[!b]
    \centering
    \includegraphics[width=\textwidth]{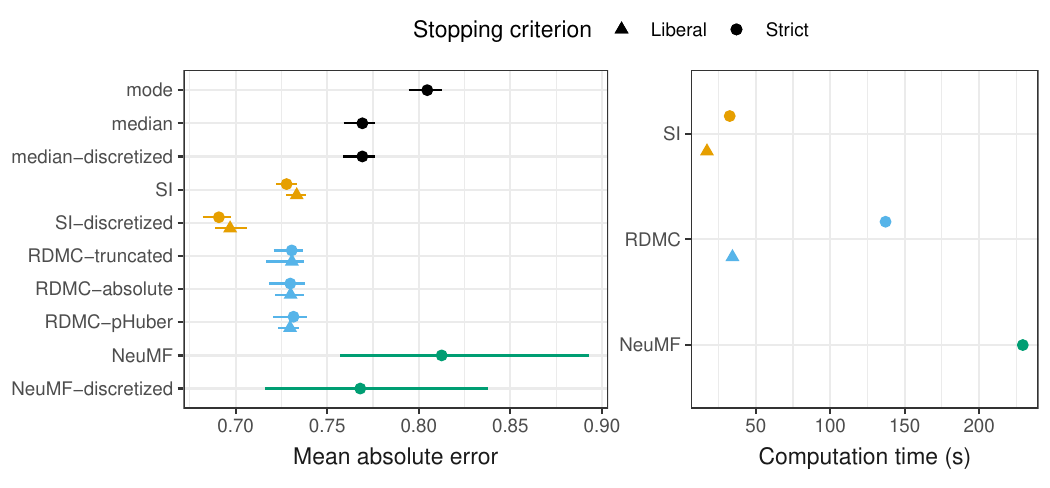}
    \caption{Baseline prediction performance in the absence of attacks (left panel) and computation time in seconds on the full dataset (right panel) for the MovieLens 100K data. Prediction performance is averaged across ten data splits (points), with the minimum and maximum also indicated (line range). Computation time is averaged over ten hyperparameter configurations and in case of RDMC also across the three considered loss functions.}
    \label{fig:movielens_no_attacks}
\end{figure}

Figure~\ref{fig:movielens_no_attacks} (left) displays the prediction performance of the different methods in terms of the average MAE on the test set across ten different data splits.
RDMC (with any of the three loss functions) and SI clearly outperform median and mode imputation with an average MAE of around 0.73, while SI-discretized presents an even smaller average MAE of 0.69. In contrast, the average MAEs of NeuMF-discretized and NeuMF are similar to those of median imputation and mode imputation, respectively, at 0.77 and 0.81.

To analyze the sensitivity of the methods to different data for training and evaluation, Figure~\ref{fig:movielens_no_attacks} (left) also includes line ranges indicating the minimum and maximum MAE across the ten data splits. RDMC, SI, as well as median and mode imputation are stable with little variation in the MAE. NeuMF, on the other hand, lacks stability with the range of the MAE being larger by an order of magnitude.

Regarding sensitivity to the stopping criterion, the prediction performance of RDMC and SI is hardly affected by applying the liberal rather than the strict criterion.
However, Figure~\ref{fig:movielens_no_attacks} (right) shows that the former yields a substantial reduction in computation time for RDMC---by a factor of four---resulting in similar computation time to SI.

\subsubsection{Results on the impact of attacks}
\label{sec:movielens-attacks}

\begin{figure}[!t]
    \centering
    \includegraphics[width=\textwidth]{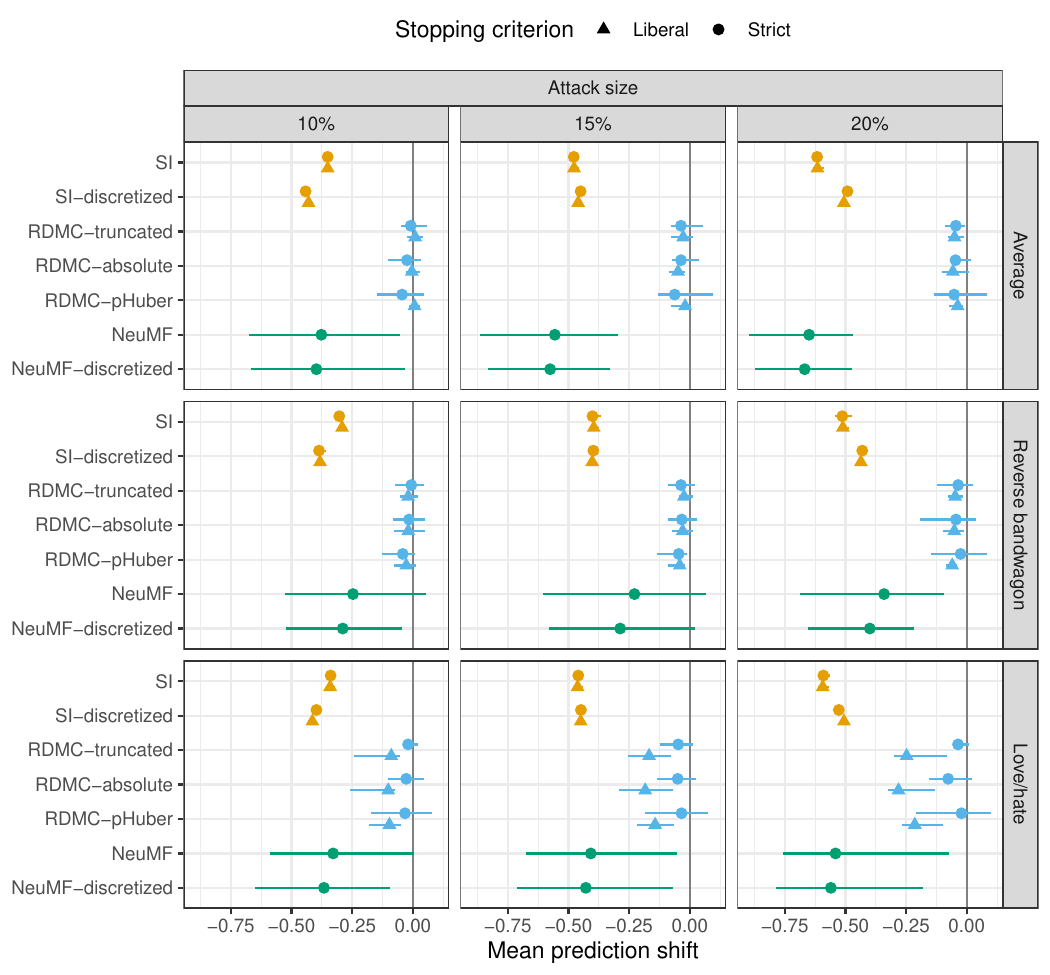}
    \caption{Robustness of predictions for the MovieLens 100K data in the presence of attacks. The rows correspond to different 
    attack schemes and the columns to different attack sizes. Results are averaged across ten data splits and attacks (points), with the minimum and maximum also indicated (line range).}
    \label{fig:movielens_attacks}
\end{figure}

We now investigate the robustness of the methods under nuke attacks.
Figure~\ref{fig:movielens_attacks} displays the average MPS across repeated analyses with ten different data splits and attacks.
All three RDMC approaches outperform both SI variants and both NeuMF variants by a considerable margin, for all attack schemes. Moreover, RDMC provides effective protection against the attacks with an average MPS close to 0 in all cases, irrespective of the loss function.
Conversely, the two SI variants and the two NeuMF variants deteriorate in terms of average MPS as the size of the attack increases.
The magnitude of the average MPS is thereby large enough that the target item may be recommended to far fewer users.

We now turn to analyzing how sensitive the results are to different data for training and evaluation.
Figure~\ref{fig:movielens_attacks} contains line ranges indicating the minimum and maximum MPS across the ten repeated analyses with different data splits and attacks. SI is highly stable, with the line ranges for the MPS often not extending beyond the size of the points. RDMC is overall relatively stable, whereas NeuMF varies greatly in terms of the MPS for all settings.

Finally, we evaluate the sensitivity with respect to the stopping criterion. SI is also highly stable in this regard. For RDMC, setting the liberal stopping criterion does not deteriorate the MPS in most settings. For the love/hate attack, however, the MPS increases slightly in magnitude, but it remains much smaller than that of SI and NeuMF.

In summary, we find RDMC to be the most reliable of the compared methods in this case study, as it is robust to attacks and its results are relatively stable to algorithmic choices. SI is highly stable but not robust, whereas NeuMF is neither stable nor robust.

\subsection{Yahoo!~Music data: The impact of missing data mechanisms}
\label{sec:Yahoo}

We use the Yahoo!~Music data to investigate the impact of missing data mechanisms on the performance of the matrix completion methods, thereby addressing Q3. For each of the two datasets (MNAR and MCAR mechanisms), we reserve 20\% of the observed ratings to evaluate out-of-sample prediction performance and use the remaining ratings for training and hyperparameter validation.

\begin{figure}[!b]
    \includegraphics[width=\textwidth]{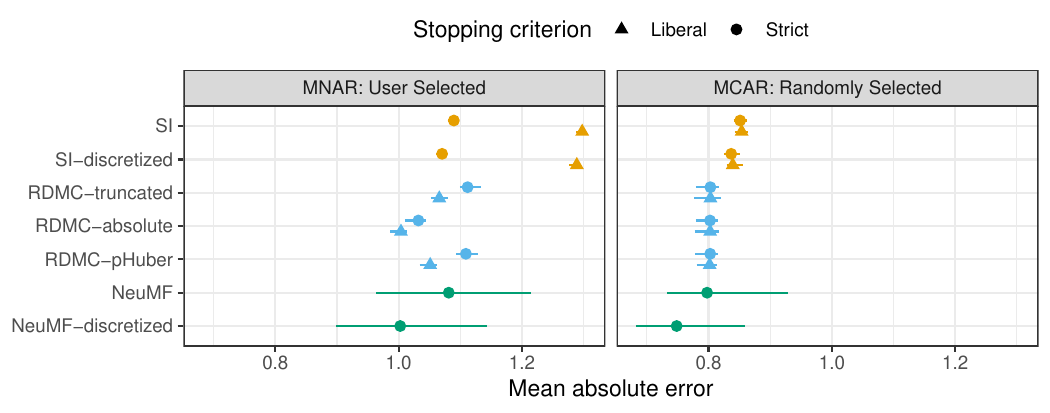}
    \caption{Prediction performance for the Yahoo!~Music datasets. The columns correspond to different missing data mechanisms.}
    \label{fig:yahoo}
\end{figure}

For both missing data mechanisms, Figure~\ref{fig:yahoo} shows the average MAE on the test set across ten different data splits. As can be expected, the average MAE of all methods is lower in case of MCAR compared to MNAR, despite the former dataset having far fewer observed ratings. Overall, RDMC outperforms the SI variants in both datasets.
For RDMC, the loss function seems to make a difference only for the MNAR data, with a slightly lower average MAE for the absolute loss function.
For SI, the discretized variant yields only a minimal improvement over the standard algorithm.
For NeuMF, the average MAE of the standard algorithm is in line with RDMC, while the discretized variant yields the lowest average MAE in particular for the MCAR data.

However, the sensitivity analysis with respect to the data used for training and evaluation sheds a different light on the latter finding. Figure~\ref{fig:yahoo} further visualizes the minimum and maximum MAE across the ten data splits via line ranges. NeuMF again lacks stability, with its MAE range extending beyond the entire spectrum for the other methods. RDMC and SI are highly stable, with the line ranges for SI often being masked by the size of the points.

Concerning sensitivity to the stopping criterion, on the other hand, the prediction performance of SI in the MNAR data deteriorates substantially when the liberal stopping criterion is applied. Yet for RDMC, there seems to be even a minor improvement with the liberal stopping criterion. At the same time, computation time of RDMC is drastically reduced with the liberal stopping criterion, making it comparable to that of SI (see Appendix~\ref{app:applications} for a discussion on computation time).

To summarize, RDMC performs well in terms of prediction under both missing data mechanisms in this case study. Among the considered methods, it is the only one that is stable with respect to both the data splits and the stopping criterion.


\section{(A blueprint for) simulation experiments}
\label{sec:simulations}

Building on the empirical insights from the case studies, we now
evaluate the methods further via simulations.
We hereby provide a blueprint for designing realistic simulation setups for matrix completion and recommender systems with explicit feedback, thus addressing scientific question Q4.
We perform 100 replications in our experiments.

\subsection{Data generation}

We simulate data of $n = 300$ user ratings on $p = 200$ items. We start by simulating latent continuous data from the low-rank matrix factorization model
$\mat{Z} = \mat{A} \mat{B}^{\top} + \mathcal{E}$, where
$\mat{A}$ is of dimension $n \times q$, $\mat{B}$ of dimension $p \times q$, and $\mat{\mathcal{E}}$ of dimension $n \times p$, with rank $q = 20$ and the elements of $\mat{A}$, $\mat{B}$, and $\mat{\mathcal{E}}$ being independent and standard normally distributed. Subsequently, we rescale $\mat{Z}^{*} = \mat{Z} / \sqrt{q+1}$ so that the elements of $\mat{Z}^{*}$ have variance 1.

To create popular items (in general higher ratings) and unpopular items (in general lower ratings), we add random mean shifts $s_{1}, \dots, s_{p}$ to the respective columns of $\mat{Z}^{*}$ before discretization into $K \in \{3, 5, 10\}$ ordinal rating categories encoded as values $\{1, \dots, K\}$. These mean shifts are randomly drawn from the interval $[-s_{\text{max}}, s_{\text{max}}]$, where $s_{\text{max}}$ depends on the number of categories and breakpoints in the discretization. Specifically, $s_{\text{max}}$ is chosen so that the corresponding mean shift results in 40\% of the ratings being expected in the maximum rating category.

For discretizing the resulting continuous data matrix in order to obtain a rating matrix $\mat{R}$, we use the following breakpoints depending on the number of rating categories:
\begin{itemize}
    \item $K = 3$: inspired by Netflix' asymmetric rating scale (`Not for me', `I like this', `Love this!'), we set the breakpoints between categories at 0 and 1.5.
    \item $K = 5$: to simulate common 1 to 5 star ratings, we set the breakpoints at $-1.5$, $-0.5$, $0.5$, and $1.5$.
    \item $K = 10$: to simulate common 1 to 10 star ratings (or, equivalently, ratings up to 5 stars but allowing for half-stars), we set the breakpoints at $-2$, $-1.5$, $-1$, $-0.5$, $0$, $0.5$, $1$, $1.5$, and $2$.
\end{itemize}

\subsection{Missing values and attacks}

We generate missing values in the rating matrix $\mat{R}$ using two different settings:
\begin{itemize}
    \item Missing not at random (MNAR): the negated mean shifts $-s_{1}, \dots, -s_{p}$ are mapped to the interval $[0.4, 0.99]$ in order to obtain the proportion of observations in each item to be replaced by missing values. For instance, the most popular item (large positive mean shift, high ratings) contains 40\% missing values, while the most unpopular item (large negative mean shift, low ratings) contains 99\% missing values. These choices are motivated by the MovieLens 100K data \citep{harper2015movielens} used in Section~\ref{sec:MovieLens}, in which the most complete item contains $38.2\%$ missing values and the most incomplete item contains more than 99\% missing values. Hence, low ratings are much more likely to be missing than high ratings.
    \item Missing completely at random (MCAR): a proportion $\gamma = 0.7$ of cells in $\mat{R}$ is randomly selected and replaced by missing values. Here, $\gamma$ is chosen so that the overall proportion of missing values is similar to the MNAR setting. Note that this setting is included for reference purposes since MCAR is unrealistic in recommender systems applications.
\end{itemize}

Beyond these missing data mechanisms, we further assess performance under adversarial manipulation. To this end, we carry out the three nuke attack schemes described in Section~\ref{sec:attacks}, but we adapt the selection of the target variable to be more in line with other aspects of the data generating process.
We first make a pre-selection of items with a mean shift larger than $0.9 \cdot s_{\text{max}}$.
Although it requires knowledge of unobserved information, this pre-selection reflects that the attacker has some prior notion of popular items. Among those, the item with the highest average observed rating is targeted.
To keep the effect of the attacks comparable across missing data settings, we again set the number of fake profiles as a proportion $\varepsilon$ of the number of observed ratings in the target item. However, we only consider $\varepsilon = 0.2$ to reduce the overall number of parameter settings.

\subsection{Results}
\label{subsubsec::results_simus_recommender}

For RDMC, we focus on the pseudo-Huber loss, as preliminary simulations indicate that results for the absolute loss and the truncated absolute loss are similar (see Appendix~\ref{app:sim-recommender}).
Figure~\ref{fig:recommender_no_attack} displays box plots of the MAE for the setting without an attack. For three rating categories, standard SI does not improve upon median and mode imputation, while standard NeuMF has by far the highest MAE.
SI-discretized yields a clear improvement in MAE, with RDMC falling in between followed by NeuMF-discretized. For five rating categories, the advanced methods considerably outperform the basic methods, with RDMC being in between the two SI variants (MNAR setting) or performing equally well as SI-discretized (MCAR setting). The NeuMF variants yield a similar median MAE as the SI variants but with much higher variability. For ten rating categories,
RDMC is in between the two NeuMF variants, but with the latter having high variability. Both SI variants have a larger MAE here and exhibit large variability in the MCAR setting.

\begin{figure}[!t]
\centering
\includegraphics[width=\textwidth]{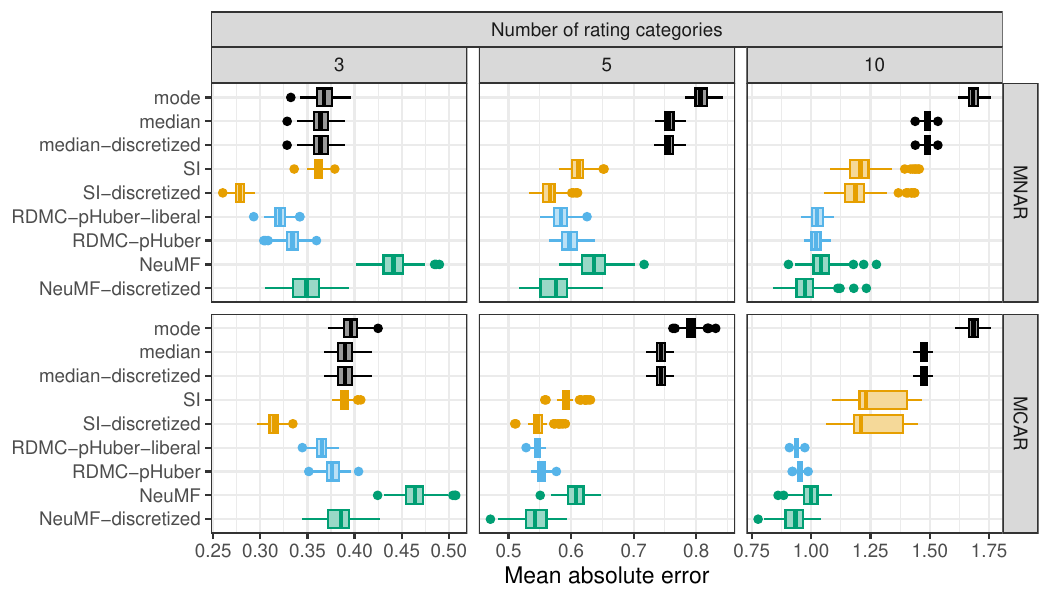}
\caption{Results for the simulated recommender system without an attack. The rows correspond to different missing data mechanisms and the columns to different numbers of rating categories.}
\label{fig:recommender_no_attack}
\end{figure}

\begin{figure}[!t]
\centering
\includegraphics[width=\textwidth]{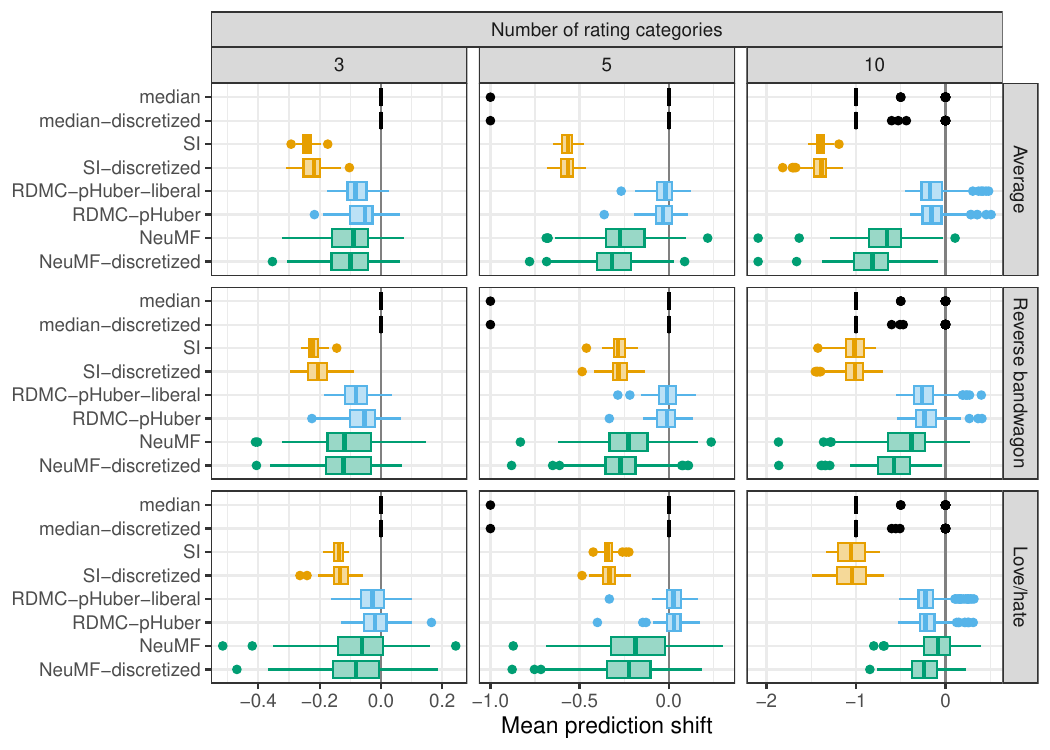}
\caption{Results for the simulated recommender system with attacks in the MNAR setting. The rows correspond to different attacks and the columns to different numbers of rating categories. Results
for mode imputation were unstable and are therefore omitted.}
\label{fig:recommender_MNAR}
\end{figure}

We now turn to robustness of the predictions under adversarial manipulation, with Figure~\ref{fig:recommender_MNAR} containing box plots of the MPS for the MNAR setting. Results for the MCAR setting are qualitatively similar and can be found in Appendix~\ref{app:sim-recommender}.
Both SI variants are strongly influenced by the attacks, with the average attack having the biggest impact.
Far fewer users may be recommended the target item---for instance, the predictions on average drop by at least one full rating category in the setting with ten categories.
The median MPS of the NeuMF variants is small in some settings but relatively large in others, with high variability across the replications.
RDMC is far more robust in the presence of attacks, with an MPS (comparatively) close to zero in all settings.

On a final note, we also included a variant of RDMC with the pseudo-Huber loss that we do not iterate until convergence. Instead, we apply a liberal stopping criterion, limiting the procedure to a maximum of 10 iterations (denoted by \emph{RDMC-pHuber-liberal} in the figures). Interestingly, the MAE in case of no attack is often smaller compared to iterating until convergence, whereas the MPS in the presence of an attack is slightly larger in absolute value in case of three rating categories.


\section{Conclusions and discussion}
\label{sec:conclusions}

In a recent review paper, \citet{leblanc2023} conclude that too few statisticians are engaged in research on recommender systems despite their potential to make consequential contributions.
This paper responds to this call by addressing several fundamental challenges that limit the practical relevance of existing methods in the academic literature.
A key challenge concerns the development of methods that remain \textit{reliable} under the complexities of real-world rating data. To this end, we introduce \textit{RDMC}, a novel matrix completion method tailored to discrete ratings, thereby addressing scientific question Q1. RDMC combines a robust loss function on the errors for the observed ratings with a discreteness constraint on the predictions and a low-rank constraint on an ancillary continuous matrix.
While we introduce RDMC with three alternative robust loss functions, their performance differed little in practice. Among them the pseudo-Huber loss is recommended overall because it typically requires fewer iterations to converge.

In case studies with MovieLens and Yahoo!~Music data, we investigate the effects of adversarial manipulation through the injection of fake user profiles (Q2) and of non-random missingness (Q3) on matrix completion performance. RDMC proved robust to attacks, produced stable rating predictions with respect to common algorithmic choices (i.e., different data splits and stopping criteria) and performed  well under both MCAR and MNAR missing data mechanisms. Overall, RDMC is unique in combining these desirable properties within a single, reliable matrix completion framework. In contrast, Soft-Impute generally yields stable predictions but remains vulnerable to fake profile attacks, whereas NeuMF is neither robust to manipulation nor stable, exhibiting considerable sensitivity to data splits.

The empirical case studies underscore the importance of evaluating recommender systems under realistic conditions. To further support this objective, we provide a blueprint (or at least a useful starting point) for designing realistic simulation experiments that incorporate key features of rating data (Q4), thereby contributing to the broader goals of enhancing generalizability, transparency, and reproducibility of recommender systems research.

Several avenues for future research remain. In particular, while RDMC provides protection against adversarial manipulation, it does not explicitly flag or identify fake/corrupted users or user–item interactions. Developing robust detection methods for such anomalies in the context of discrete rating-scale data is challenging but presents an interesting direction for future research \citep[cf.][]{welz2024robust, welz2025polychoric}.

Furthermore, RDMC shows promise for applicability across diverse domains.
For instance, empirical research in social and behavioral sciences relies heavily on rating-scale surveys, but listwise deletion remains the predominant method of handling missing data \citep[e.g.,][]{peng2023handling}. Preliminary simulations suggest that RDMC represents a notable advancement in robust imputation of such data (see Appendix~\ref{app:survey}), even in the presence of inattentive respondents or bot respondents; a critical issue in online surveys.
Our approach also holds potential for contexts involving strategic missing data situations \citep{zhang2020predictive}, such as financial reporting, college admissions, job applications, and marketing, where individuals may intentionally withhold information to achieve favorable outcomes.
Finally, recent research has shown a notable rise in matrix completion methods for panel data \citep[e.g.][]{athey2021matrix, bai2021matrix, chen2024dynamic, choi2024matrix}, providing a promising direction for future adaptation of our robust methodology.

\section*{Data availability statement}\label{data-availability-statement}

The MovieLens 100K data are publicly available from \url{https://grouplens.org/datasets/movielens/100k/}.
The Yahoo!~Music datasets `R3 – Yahoo!~Music ratings for User Selected and Randomly Selected Songs, Version 1.0' were obtained on September 9, 2024, from the Yahoo!~Webscope data sharing program at \url{https://webscope.sandbox.yahoo.com/catalog.php?datatype=r} under a research-use agreement.

\section*{AI statement}\label{AI-statement}

An AI tool, specifically ChatGPT-5.5, was used to assist with code for hyperparameter tuning of NeuMF in the software package \pkg{RecBole} \citep{zhao2021recbole}, as well as with modifying its source code to allow for explicit feedback. Thorough checks and necessary manual adjustments were made by the authors.


\bibliography{references.bib}


\clearpage
\appendix

\renewcommand\thefigure{\thesection.\arabic{figure}}
\renewcommand\thetable{\thesection.\arabic{table}}
\renewcommand\thealgorithm{\thesection.\arabic{algorithm}}
\renewcommand{\theequation}{\thesection.\arabic{equation}}


\setcounter{algorithm}{0}

\section{Pseudo-code of the algorithm}
\label{app:algorithm}

\begin{algorithm}[!h]
\caption{Robust matrix completion for discrete rating-scale data} \label{algorithm-rdmc}
\hspace*{\algorithmicindent}
\textbf{Input} Incomplete rating matrix $\bf R$, loss function $\rho(\cdot)$, regularization parameter $\lambda$, regularization parameter $\mu$, update factor $\delta$, convergence threshold $\varepsilon_{\text{tol}}$, maximum number of iterations~$t_{\text{max}}$ \\
\hspace*{\algorithmicindent} \textbf{Output} Complete rating matrix $\widehat{\bf R}$
\begin{algorithmic}[1]
\State $M_{j} \leftarrow \median_{i: (i,j) \in \Omega} \{R_{ij}\}$ for $j=1,\ldots, p$
\State $X_{ij}  \leftarrow R_{ij} - M_j$ for $i=1, \ldots, n$ and $j=1,\ldots, p$
\State ${\bf L}^{(0)} \leftarrow  P_\Omega({\bf X})$
\State ${\boldsymbol\Theta}^{(0)} \leftarrow  {\bf 0}$
\State $\text{Loss}^{(0)} \leftarrow \infty$
\State $\text{converged} \leftarrow \texttt{FALSE}$
\State $t \leftarrow 1$

\While{$\neg$\text{converged} $\And$ $t \leq t_\text{max}$}

\State $ { {\bf Z}^{(t)} \leftarrow  \displaystyle \argmin_{\mat{Z}}}
{\textstyle \
\frac{1}{2} \| (\mat{L}^{(t-1)} + \frac{1}{\mu} \mat{\Theta}^{(t-1)}) - \mat{Z} \|_{F}^{2} +
\frac{\lambda}{\mu} \| \mat{Z} \|_{*}
}$ \Comment{using Equation \eqref{eq:Zupdate}}

\State ${\bf L}^{(t)} \leftarrow  \displaystyle \argmin_{{\bf L}}    \| P_{\Omega}(\mat{X}) - P_{\Omega}(\mat{L}) \|_{\rho} + \frac{\mu}{2} \| \mat{L} - \mat{Z}^{(t)} + \frac{1}{\mu} \mat{\Theta}^{(t-1)} \|_{F}^{2}$

    \hspace{1cm} subject to  $L_{ij} \in \mathcal{C}_{j}$ \Comment{using Equation \eqref{eq:updateL}}

    \State ${\boldsymbol\Theta}^{(t)} \leftarrow {\boldsymbol\Theta}^{(t-1)} + \mu({\bf L}^{(t)} - {\bf Z}^{(t)} ) $
    \State $\mu \leftarrow \delta\mu$

    \State  $\text{Loss}^{(t)} \leftarrow \| P_\Omega({\bf X})-P_\Omega({\bf L}^{(t)})\|_{\rho} +\lambda\|{\bf Z}^{(t)}\|_* + \langle\boldsymbol\Theta^{(t)}, {\bf L}^{(t)}-{\bf Z}^{(t)}\rangle_F +\frac{\mu}{2} \|{\bf L}^{(t)}-{\bf Z}^{(t)}\|_F^2 $

    \If{$t > 1$}
		\State $\text{converged} \leftarrow |(\text{Loss}^{(t)} - \text{Loss}^{(t-1)})/\text{Loss}^{(t-1)}| \leq \varepsilon_{\text{tol}}$
    \EndIf

    \State $t \leftarrow t+1$
\EndWhile
\State $\widehat{R}_{ij} \leftarrow
\begin{cases} \textstyle
R_{ij}               & \text{ if } (i,j) \in \Omega \\
{L}^{(t)}_{ij} + M_j & \text{ otherwise}
\end{cases}
$
\end{algorithmic}
\end{algorithm}


\setcounter{equation}{0}

\section{Benchmark methods}
\label{app:benchmarks}

\paragraph{Median and Mode Imputation.}
Basic benchmark methods are median imputation (denoted by \emph{median}), a discretized variant thereof (\emph{median-discretized}; in case the median of a column falls in between rating categories, the predictions are randomly sampled from the corresponding two categories), and mode imputation (denoted by \emph{mode}; in case of a column with multiple modes, one of them is selected at random for each missing cell).

\paragraph{Soft-Impute.}
Soft-Impute (\emph{SI}) \citep{mazumder2010spectral, hastie2015matrix} is a widely-used statistical method for matrix completion based on nuclear norm regularization.
Since it yields continuous predictions, we also introduce a variant (\mbox{\emph{SI-discretized}}) that discretizes the obtained predictions to the given rating scale $\{1, \dots, K\}$ via the mapping function
\begin{equation} \label{eq:discretization}
m(y) = \min(\max([y], 1), K),
\end{equation}
where $[\cdot]$ denotes rounding to the nearest integer. We use the SVD-based algorithm due to its theoretical convergence guarantees and set the same convergence threshold as for RDMC.
The regularization parameter~$\lambda$ is selected using the same procedure as for RDMC, except that the candidate values are scaled by the largest singular value based on the mean-centered training data.

\paragraph{Neural Matrix Factorization.}
Neural Matrix Factorization (\emph{NeuMF}) \citep{he2017neural} combines the linearity of matrix factorization and the nonlinearity of deep neural networks for modeling latent structures in user-item interactions. It is considered state-of-the-art for deep learning in recommender systems \citep{raza2026comprehensive}. While this method is mainly used for implicit feedback---as implemented in the software package \pkg{RecBole} \citep{zhao2021recbole} with the sigmoid activation function in the output layer and the cross-entropy loss function---it can also be applied to explicit feedback in the form of ratings \citep[cf.][]{zhang_deep_2019}.
To this end, we modified the implementation from \pkg{RecBole} to allow for typical architecture choices for explicit feedback, namely the identity activation function in the output layer and the mean-squared-error (MSE) loss function. Furthermore, we again introduce a variant (\emph{NeuMF-discretized}) with the post-hoc discretization step from~\eqref{eq:discretization} for the predictions. As \pkg{RecBole} does not support repeated holdout validation, we use a single validation set of the same size as for RDMC. Inspired by \citet{chen2020efficient}, we use the following candidate values for hyperparameter tuning: $\{0.1, 0.3, 0.5, 0.7, 0.9, 1\}$ for the dropout rate, $\{0.005, 0.01, 0.02, 0.05\}$ for the learning rate, $\{8, 16, 32, 64\}$ for the latent dimension of the matrix factorization component, $\{ \{16,8\}, \{32,16\}, \{64,32\}, \{128,64\} \}$ for the latent dimensions in the deep neural network component, and $\{128, 256, 512, 1024\}$ for the training batch size. As there are thousands of possible hyperparameter configurations, we apply random sampling from this hyperparameter space with \pkg{RecBole}'s default settings of evaluating a maximum of 100 random combinations and early stopping if there is no improvement for 10 consecutive random combinations.


\setcounter{figure}{0}

\section{Additional results for the empirical case studies}
\label{app:applications}

\begin{figure}[!t]
  \includegraphics[width=\textwidth]{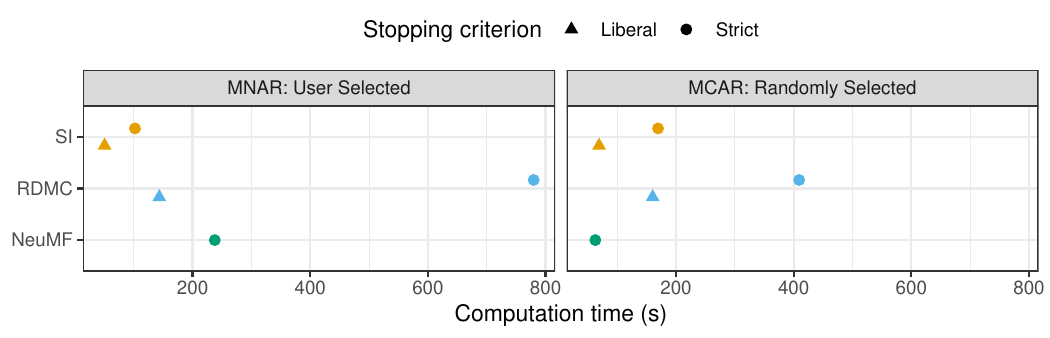
    }
    \caption{Computation time in seconds for the Yahoo!~Music datasets, averaged over ten hyperparameter configurations and in case of RDMC also across the three considered loss functions. The columns correspond to different missing data mechanisms.}
    \label{fig:yahoo_time}
\end{figure}

Figure~\ref{fig:yahoo_time} displays the computation time of the matrix completion methods for the Yahoo!~Music datasets from Section~\ref{sec:Yahoo}.
While NeuMF had the longest computation time per hyperparameter combination on the MovieLens data (see the right panel of Figure~\ref{fig:movielens_no_attacks}), it is much faster here in comparison. On the MNAR data, NeuMF is closer to RDMC with the liberal criterion than with the strict criterion in terms of computation time per hyperparameter, and it is even the fastest overall for the MCAR data. Nevertheless, while we always use only 10 candidate values for $\lambda$ in RDMC and SI, NeuMF typically requires evaluating a much larger number of hyperparameter configurations.

\clearpage


\setcounter{figure}{0}

\section{Additional simulation results}
\label{app:sim-recommender}

\begin{figure}[!b]
\centering
\includegraphics[width=0.31\textwidth]{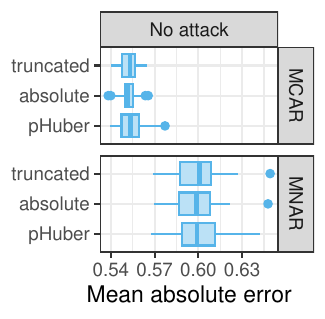}
\includegraphics[width=0.68\textwidth]{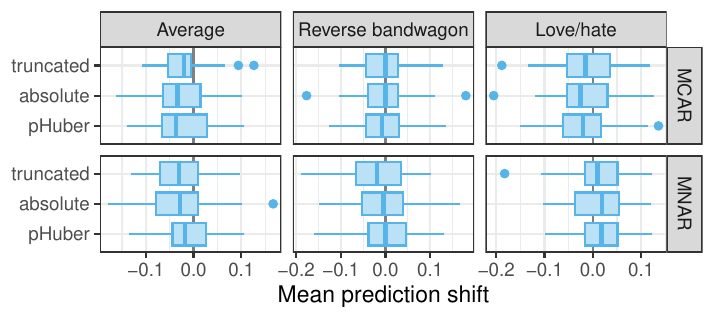}
\caption{Preliminary results for RDMC with different loss functions for the simulated recommender system with five rating categories. The left plot shows the mean absolute error in the setting without an attack, while the right plot displays the mean prediction shift for different attacks. The rows correspond to different missing data mechanisms.}
\label{fig:recommender_RDMC}
\end{figure}

\begin{figure}
\centering
\includegraphics[width=\textwidth]{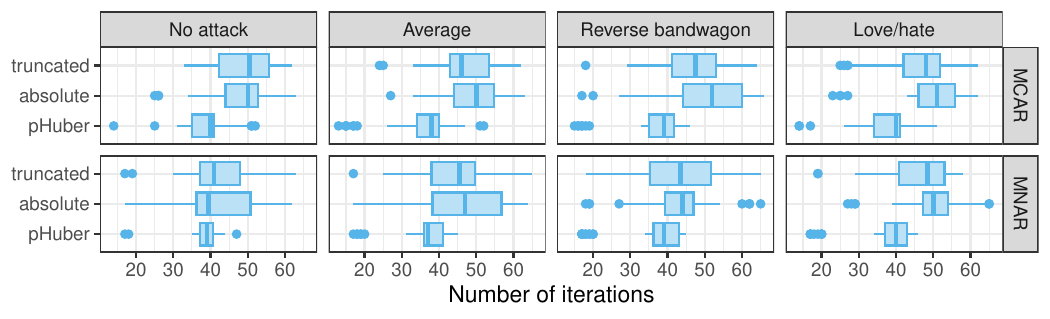}
\caption{Number of iterations for RDMC in preliminary simulations of a recommender system with five rating categories. The rows correspond to different missing data mechanisms and the columns to different attacks.}
\label{fig:recommender_RDMC_iterations}
\end{figure}

Figure~\ref{fig:recommender_RDMC} contains preliminary results for RDMC with different loss functions from 50 replications of the simulated recommender system with five rating categories, where the left plot displays the MAE for the setting without an attack and the right plot presents the MPS for different attack schemes. All three loss functions yield similar results, while the pseudo-Huber loss typically requires fewer iterations to converge (see Figure~\ref{fig:recommender_RDMC_iterations}).

\begin{figure}
\centering
\includegraphics[width=\textwidth]{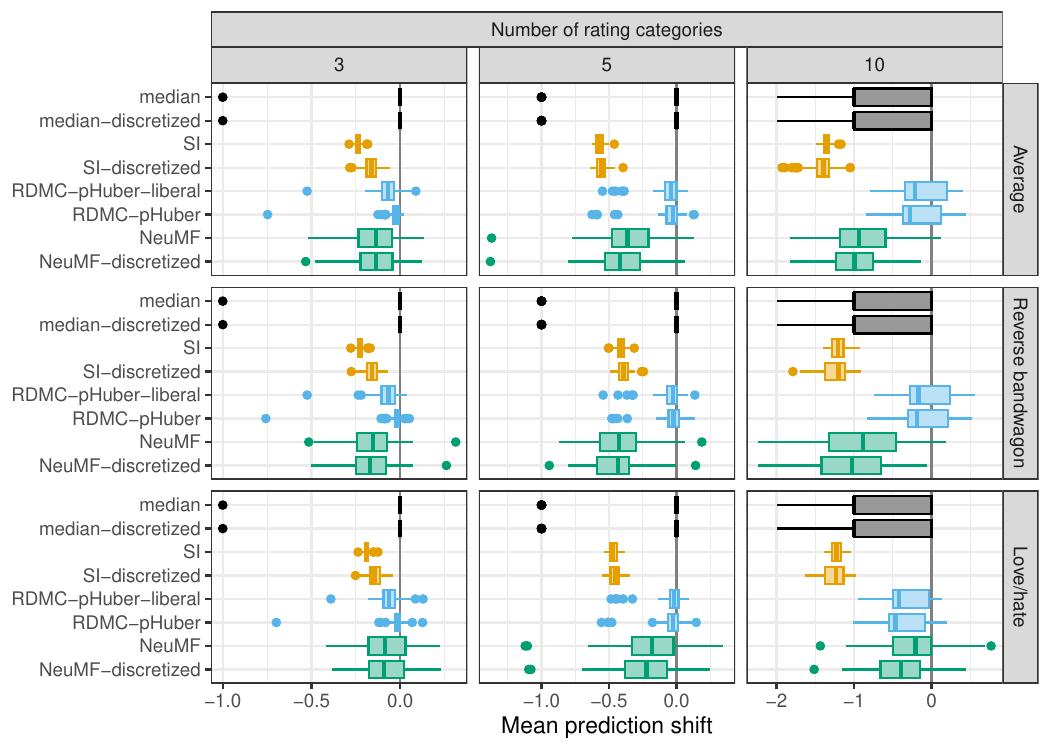}
\caption{Results for the simulated recommender system with attacks in the MCAR setting. The rows correspond to different attacks and the columns to different numbers of rating categories. Results for mode imputation were unstable and are therefore omitted.}
\label{fig:recommender_MCAR}
\end{figure}

Figure~\ref{fig:recommender_MCAR} shows the MPS under different attacks for the simulated recommender system in the MCAR setting. The results are qualitatively similar to those of the MNAR setting, but a notable difference is that there is a small number of instances where RDMC yields a relatively large negative prediction shift. A possible explanation why this occurs in the MCAR setting but not in the MNAR setting is that in the former, all variables including the target variable contain equally sparse information. In practice, however, this is a rather unrealistic setting.

\setcounter{figure}{0}

\section{Rating-scale surveys}
\label{app:survey}

\subsection{Introduction}

Another relevant application of rating-scale data, although rarely considered in the literature on matrix completion, are surveys in the social and behavioral sciences. In such surveys, researchers measure each latent construct of interest (e.g., personality traits) by multiple rating-scale items (e.g., respondents may be asked how accurately certain adjectives describe their personality, with response categories ranging from 1 = ``very accurate'' to 5 = ``very accurate''; \citealp{arias2020}). Missing values are commonly occurring due to nonresponse in parts of the survey, with nonresponse often being missing not at random (MNAR). For instance, participants may abandon the survey yielding higher probability of missingness for later survey items, or they may not respond to items they find, e.g., confusing or inappropriate. In addition, \emph{careless responding} (i.e., responses not based on item content due to inattention or misunderstanding; e.g., \citealp{ward2023}) or \emph{bot responding} \citep[e.g.,][]{storozuk2020bot} are common occurrences in online surveys, and a prevalence as low as 5\% can invalidate research findings \citep[e.g.,][]{crede2010, arias2020, welz2024maxbias}.

In some fields, removing participants with missing responses is still the most common method of handling missing responses in survey data \citep[e.g.,][]{peng2023handling}. Yet this constitutes a loss of valuable information (the observed responses of the removed participants) and generally results in biased analyses \citep[e.g.,][]{little2019statistical}. Furthermore, the common presence of careless respondents or bot respondents requires imputation methods designed for discrete rating-scale data. We therefore investigate the use of the proposed matrix completion method RDMC in  preliminary simulations motivated by surveys in the social and behavioral sciences.

\subsection{Simulations}
\label{sec:sim-survey}

\subsubsection{Data generation}

We simulate rating-scale responses of $n = 300$ participants to a survey on $q = 10$ latent constructs, with each construct being measured by $r \in \{4, 8\}$ items such that the total number of items is $p \in \{40, 80\}$.
We thereby simulate the underlying latent sentiments $\mat{Z}$ from a multivariate normal distribution $N(\vect{0}, \mat{\Sigma})$, where $\mat{\Sigma}$ follows a block-Toeplitz structure with 1 on the diagonal and the remaining elements being given by $\rho_{kl} = 0.6^{|k-l|+1}$ with $k = 1, \dots, q$ and $l = 1, \dots, q$ denoting the row and column indices of the blocks in~$\mat{\Sigma}$.

In survey-based research, skewness towards one side of the rating scale is common, for instance, in organizational and management research \citep[cf.][]{becker2019transformations,alfons2022mediation}.
Hence, we generate items with skewed distributions by adding random mean shifts to the columns of $\mat{Z}$---with all items within the same construct receiving the same mean shift---before discretization into $K \in \{5, 7, 9\}$ answer categories interpreted as values $\{1, \dots, K\}$.
The mean shifts are drawn from the interval $[0, s_{\text{max}}]$, where $s_{\text{max}}$ is chosen to result in 40\% of the responses being expected in the highest answer category.

The resulting continuous data matrix is then discretized using equispaced breakpoints $-K/2 + 1, -K/2 + 2, \dots, K/2 -2, K/2 -1$. To simulate reverse-keyed items, which are commonly used in practice to help detect response inattention, half the items of each construct are reversed by reassigning answer category 1 to $K$, 2 to $K-1$, and so on.
This implies that the data contains both left-skewed and right-skewed items. As is common in the behavioral sciences, we randomize the order of the items in the survey (with the same order being used for all participants), resulting in the final rating-scale data matrix~$\mat{R}$.

\subsubsection{Missing values and careless responding}

While online data collection tools can be set up to require participants to respond to individual questions in order to progress in the survey, participants may abandon the survey altogether.
We therefore inject missing values into the data matrix $\mat{R}$ by simulating respondents who stop responding at some point in the survey. We set the proportion of respondents who abandon the survey to $\gamma \in \{0.2, 0.4, 0.6\}$. For each of those respondents, we randomly select the item at which they stop responding, with all responses from this item onwards being replaced by missing values. We emphasize that here $\gamma$ does not indicate the proportion of missing cells in $\mat{R}$, but the number of rows that contain missing values.

To simulate careless respondents, a proportion $\varepsilon \in \{0, 0.1, 0.2\}$ of the rows in $\mat{R}$ are replaced with careless respondents who randomly select from the two extreme answer categories $\{1, K\}$. Although this is a common careless response style \citep[e.g.,][]{baumgartner2001, alfons2024open}, it may not be realistic that all careless respondents behave in this way. Yet we chose this careless response style because the high leverage of the responses should be particularly influential for matrix completion methods.
Note that our simulation allows for the situation that a careless respondent abandons the survey such that a certain row may contain both careless responses and missing values.

\subsubsection{Evaluation criteria}

We compare the methods by computing the mean absolute error (MAE) over the predictions of missing cells as in~\eqref{eq:MAE}, but excluding missing cells in rows corresponding to careless respondents. The reason for excluding the latter is that in a practical setting, careless respondents should be handled by detection and removal, or better by applying robust methods that downweight such outlying observations \citep[e.g.,][]{alfons2024open}.

\subsubsection{Results}

\begin{figure}[!t]
\centering
\includegraphics[width=\textwidth]{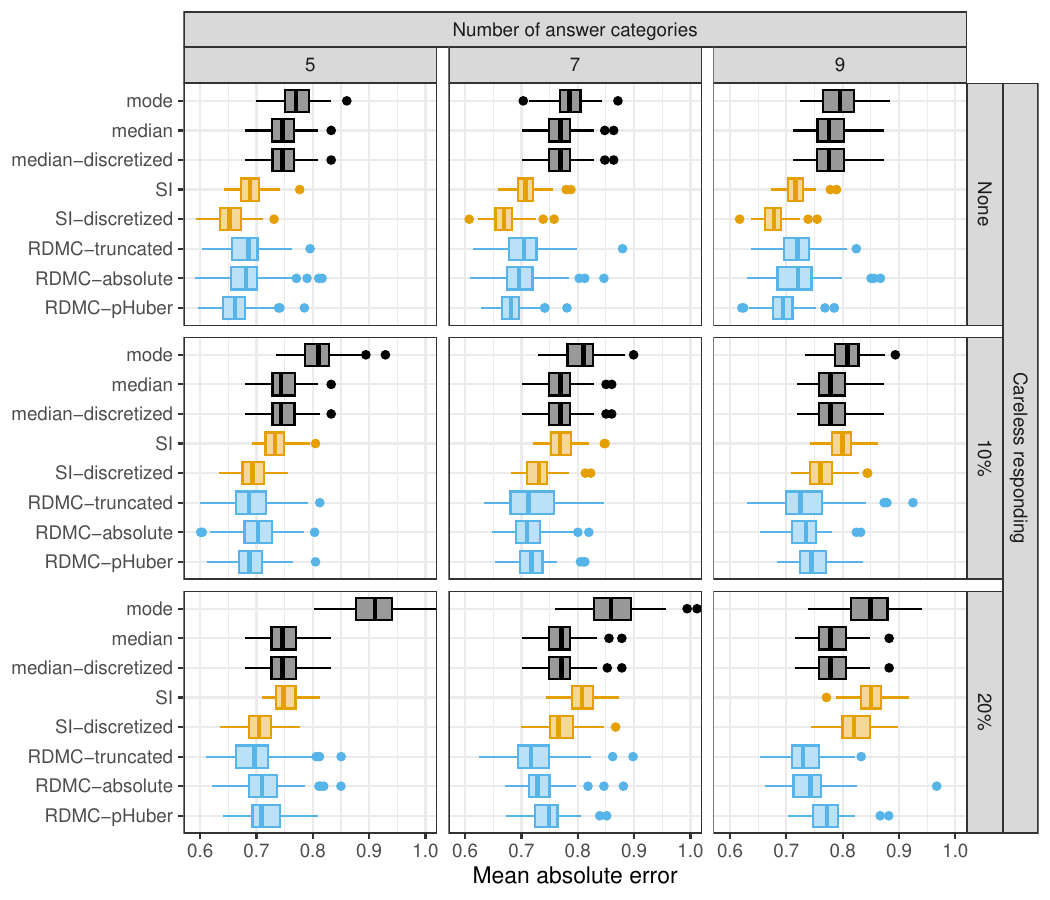}
\caption{Results for the simulated survey with $p = 40$ variables and 20\% of respondents who abandon the survey. The rows correspond to different number of answer categories and the columns to different proportions of careless respondents.}
\label{fig:survey_40_20}
\end{figure}

\begin{figure}[!t]
\centering
\includegraphics[width=\textwidth]{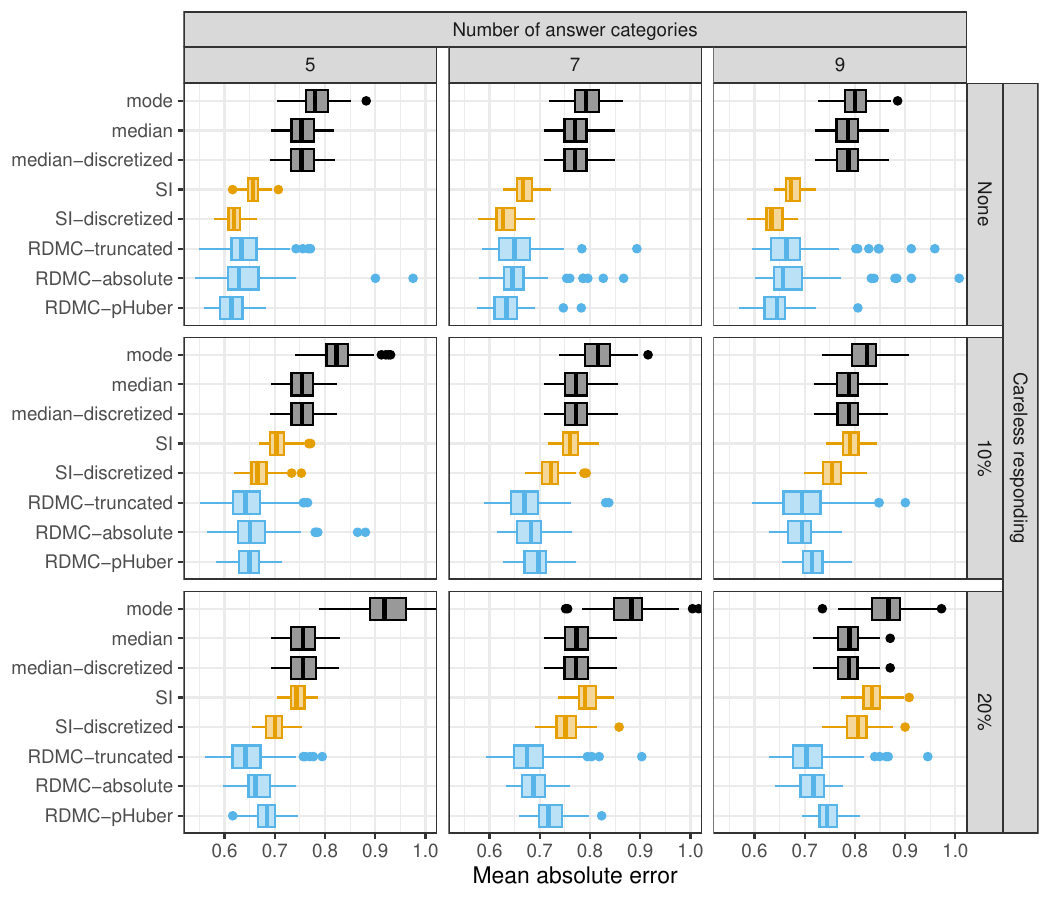}
\caption{Results for the simulated survey with $p = 80$ variables and 20\% of respondents who abandon the survey. The rows correspond to different number of answer categories and the columns to different proportions of careless respondents.}
\label{fig:survey_80_20}
\end{figure}

\begin{figure}[!t]
\centering
\includegraphics[width=\textwidth]{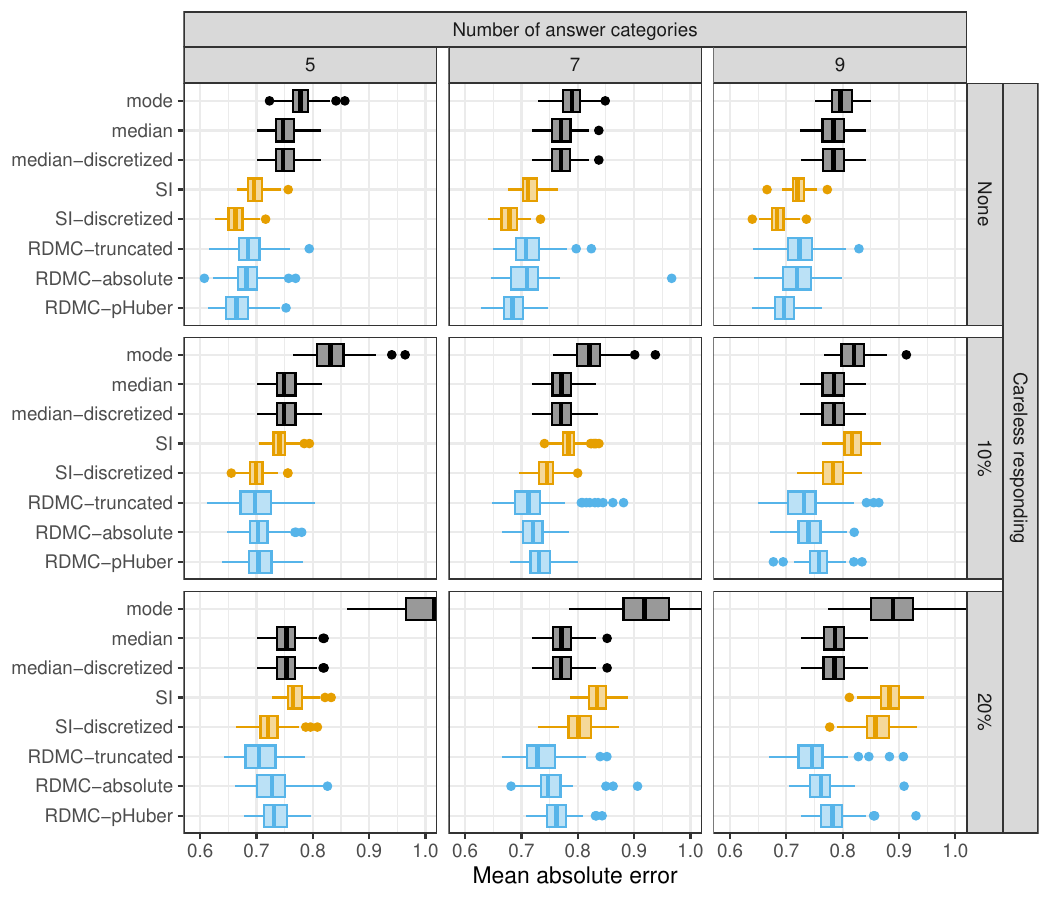}
\caption{Results for the simulated survey with $p = 40$ variables and 40\% of respondents who abandon the survey. The rows correspond to different number of answer categories and the columns to different proportions of careless respondents.}
\label{fig:survey_40_40}
\end{figure}

\begin{figure}[!t]
\centering
\includegraphics[width=\textwidth]{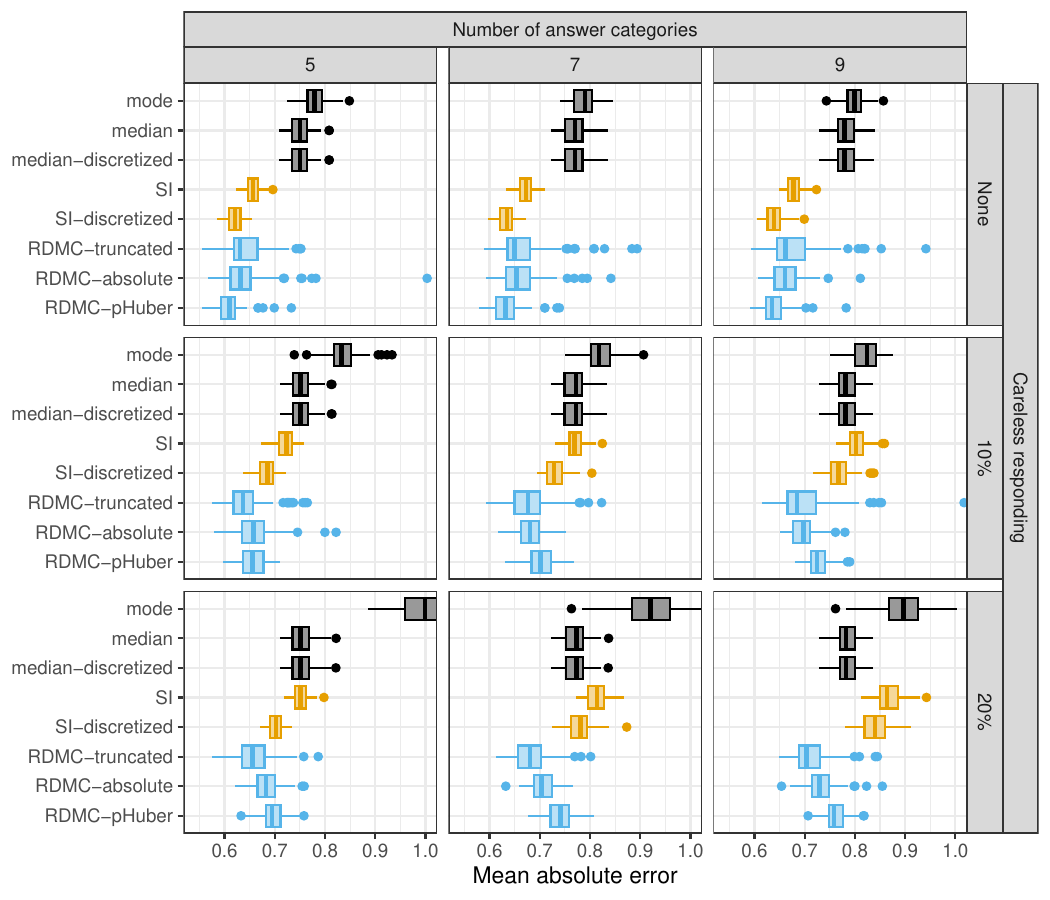}
\caption{Results for the simulated survey with $p = 80$ variables and 40\% of respondents who abandon the survey. The rows correspond to different number of answer categories and the columns to different proportions of careless respondents.}
\label{fig:survey_80_40}
\end{figure}

\begin{figure}[!t]
\centering
\includegraphics[width=\textwidth]{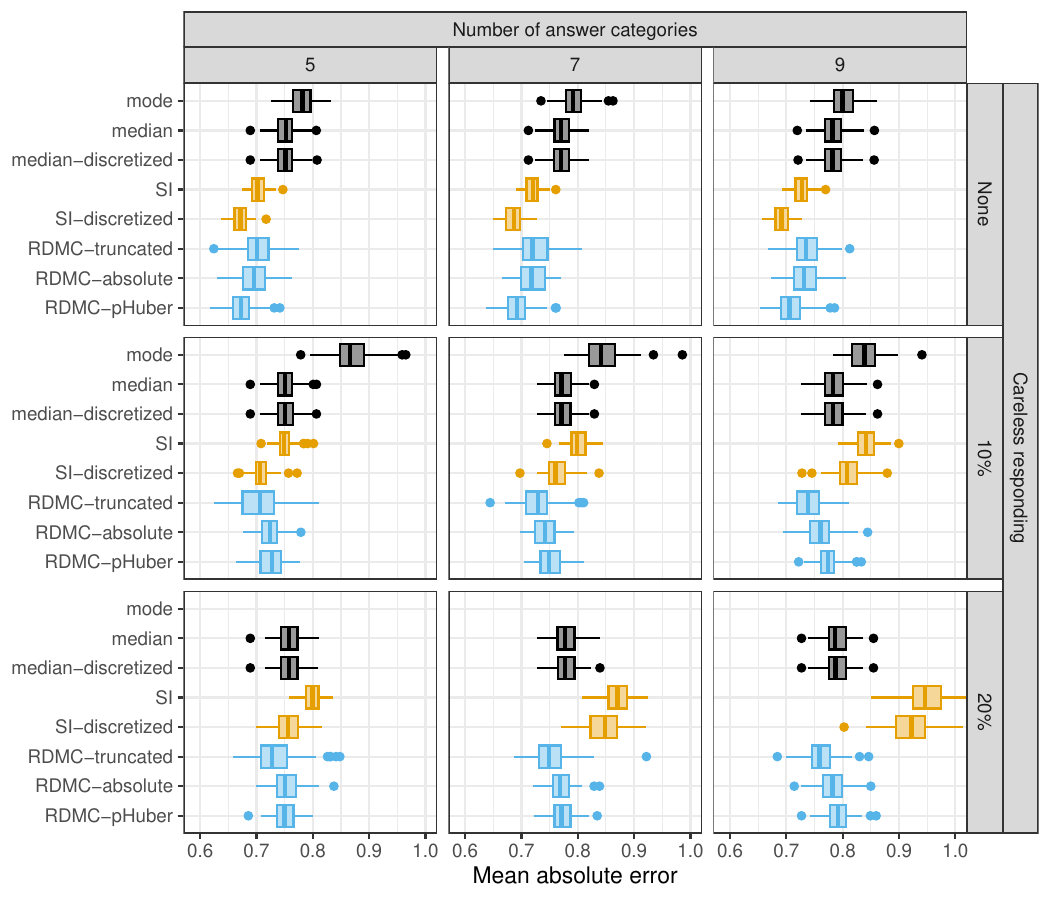}
\caption{Results for the simulated survey with $p = 40$ variables and 60\% of respondents who abandon the survey. The rows correspond to different number of answer categories and the columns to different proportions of careless respondents.}
\label{fig:survey_40_60}
\end{figure}

\begin{figure}[!t]
\centering
\includegraphics[width=\textwidth]{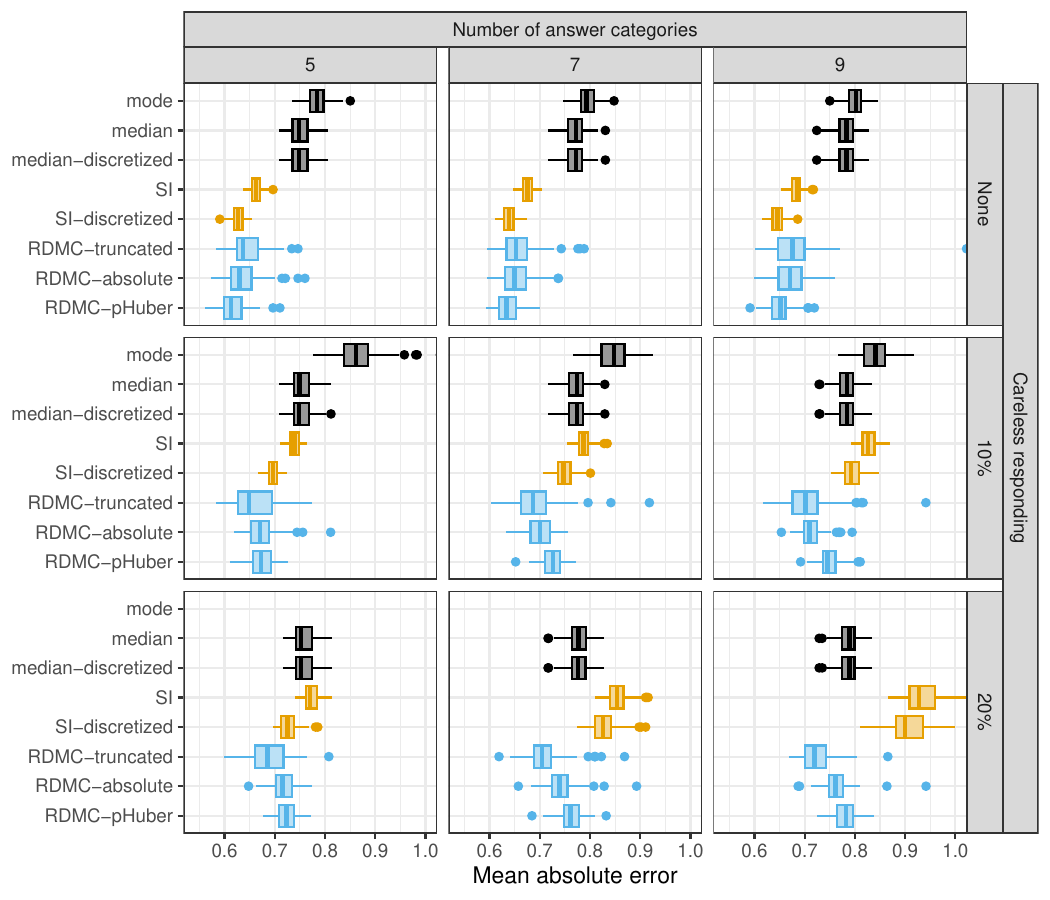}
\caption{Results for the simulated survey with $p = 80$ variables and 60\% of respondents who abandon the survey. The rows correspond to different number of answer categories and the columns to different proportions of careless respondents.}
\label{fig:survey_80_60}
\end{figure}

Since survey scales are usually validated using linear factor models, we limit the comparison to linear matrix factorization methods and basic benchmark methods.
We focus on the results from Figures~\ref{fig:survey_40_20} and~\ref{fig:survey_80_20} for the settings with 20\% of respondents who abandon the survey, as the results for 40\% and 60\% are similar (see Figures~\ref{fig:survey_40_40}--\ref{fig:survey_80_60}).

Box plots for the MAE in the simulated survey with $p=40$ items are shown in Figure~\ref{fig:survey_40_20}.
In general, differences between the methods are smaller than in the simulations for recommender systems. On the other hand, we now see more differences among RDMC with different loss functions. Most notably, RDMC with the pseudo-Huber function remains close to SI-discretized in the absence of careless responding, while RDMC with the truncated absolute loss yields the best results for 20\% careless respondents. In addition, with a higher number of answer categories, robustness benefits of RDMC over SI become more apparent. For instance, with 10\% of careless respondents, RDMC and SI-discretized perform similarly for five answer categories, whereas all three variants of RDMC outperform SI-discretized in case of seven or nine answer categories.

Increasing the number of items to $p=80$, Figure~\ref{fig:survey_80_20} contains the corresponding box plots of the MAE. We now observe bigger differences between the methods. For all settings regarding the number of answer categories, the MAE of RDMC remains in between that of the two SI approaches in case of no careless responding, while RDMC yields better performance in the presence of careless respondents. As previously, the pseudo-Huber loss is preferable in the absence of careless responding (with an MAE close to that of SI-discretized), while the truncated absolute loss is the most robust for larger prevalence of careless respondents.

Finally, we included variants of RDMC with the three loss functions for which we perform at most 10 iterations. Here, the MAE is at best similar but in most cases slightly higher compared to iterating until convergence.
We thus omit these variants from the figures.

In summary, the choice of loss function seems to play a more important role in a survey context than in our simulations of recommender systems (see Section~\ref{sec:simulations}). The pseudo-Huber loss is recommended if one expects relatively few careless respondents, whereas the truncated absolute loss is recommended if one expects a sizable number of careless respondents.

\end{document}